%% file: main.tex
\documentclass[11pt,a4paper]{article}
\usepackage[hyperref]{naaclhlt2019}
\usepackage{times}
\usepackage{latexsym}
\usepackage{graphicx}
\usepackage{multirow}
\usepackage{booktabs}
\usepackage{url}
\usepackage{xspace}
\usepackage[most]{tcolorbox}
\usepackage{makecell}
\usepackage{amssymb}  
\usepackage{tabularx}
\usepackage{booktabs}
\usepackage{makecell}
\usepackage{adjustbox}
\usepackage{float}
\usepackage{listings}
\aclfinalcopy 

\newcommand{\sys}{\mbox{\textsc{Chain-of-Code Collapse}}\xspace}

\title{\sys: Reasoning Failures in LLMs via Adversarial Prompting in Code Generation}

\author{Jaechul Roh, Varun Gandhi, Shivani Anilkumar, Arin Garg \\
  {\small \tt \{jroh, vgandhi, sanilkumar, aringarg\}@umass.edu} \\
  \\}

\date{}

\begin{document}
\maketitle

% \textcolor{purple}{
% \textbf{TODO LIST:}
% \begin{itemize}
%     \item Figure 1 (Overview) (\checkmark)
%     \item Appendix: Add all prompt templates (\checkmark)
%     \item Abstract (modify based on results finding)
%     \item Introduction
%     \item Experimental Setting: benchmark, model, evaluation metrics (\checkmark)
%     \item Experimental Results: Result, Analysis  (\checkmark)
%     \item Ablation Study (\checkmark)
%     \item Discussion \& Limitation (\checkmark)
% \end{itemize}
% }

\begin{abstract}
Large Language Models (LLMs) have achieved remarkable success in tasks requiring complex reasoning, such as code generation, mathematical problem solving, and algorithmic synthesis --- especially when aided by reasoning tokens and Chain-of-Thought prompting. Yet, a core question remains: do these models truly reason, or do they merely exploit shallow statistical patterns? In this paper, we introduce \sys, where we systematically investigate the \textit{robustness} of reasoning LLMs by introducing a suite of semantically faithful yet adversarially structured prompt perturbations. Our evaluation --- spanning 700 perturbed code generations derived from LeetCode-style problems --- applies transformations such as storytelling reframing, irrelevant constraint injection, example reordering, and numeric perturbation. We observe that while certain modifications severely degrade performance (with accuracy drops up to -42.1\%), others surprisingly \textit{improve} model accuracy by up to 35.3\%, suggesting sensitivity not only to semantics but also to surface-level prompt dynamics. These findings expose the fragility and unpredictability of current reasoning systems, underscoring the need for more principles approaches to reasoning alignments and prompting robustness. We release our perturbation datasets and evaluation framework~\footnote{\url{https://github.com/jrohsc/Chain-of-Code-Collapse}} to promote further research in trustworthy and resilient LLM reasoning.
\end{abstract}

\input{sections/intro}

\input{sections/related_works}
\input{sections/approach}
\input{sections/data}

\input{sections/experiments}

\bibliographystyle{apalike}
\footnotesize
\bibliography{yourbib}
\appendix
\input{sections/appendix}

\end{document}

%% file: sections/intro.tex
\begin{figure*}[ht]
    \centering
    \includegraphics[width=\linewidth]{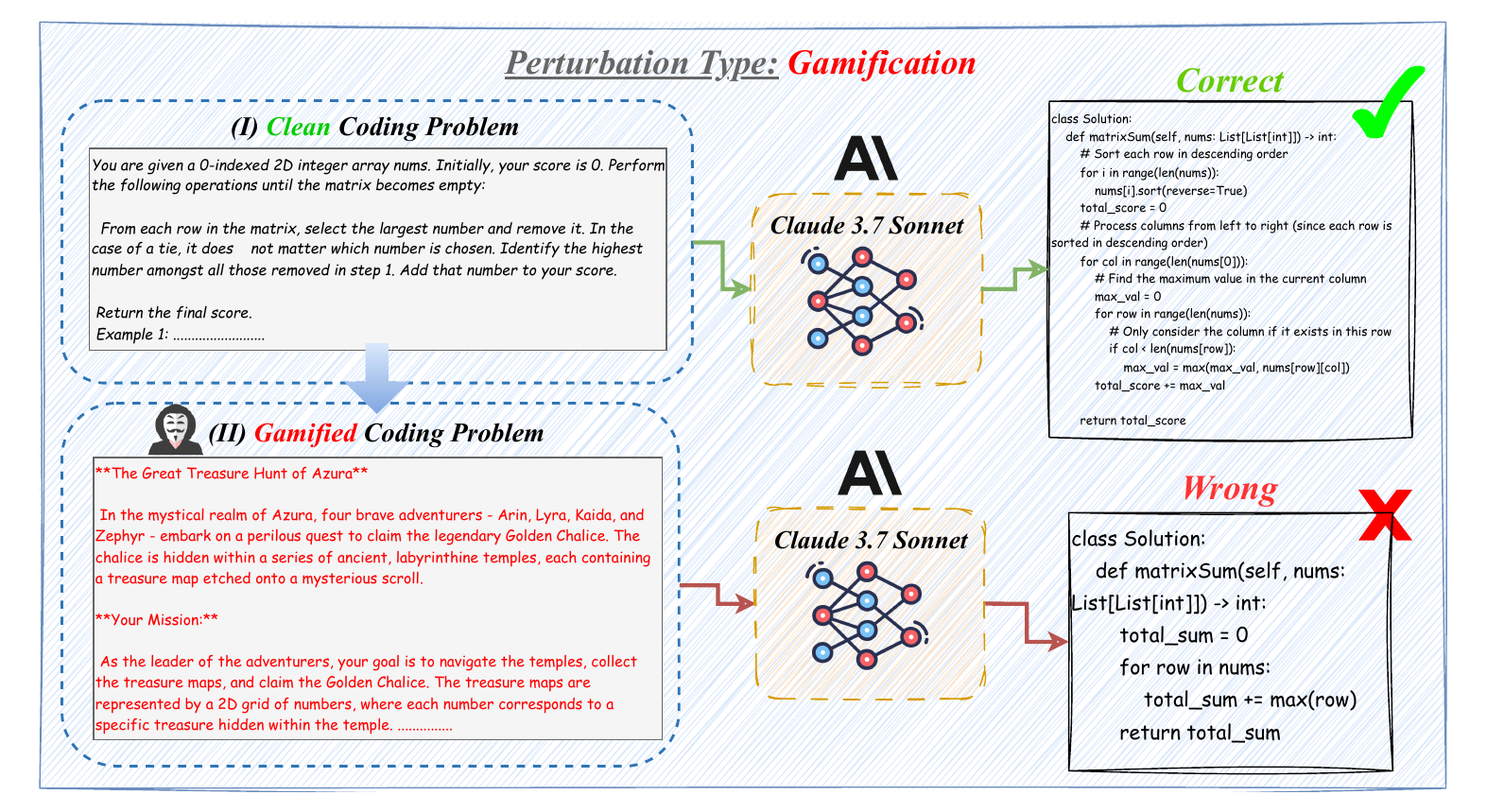}
    \caption{Overview of \sys}
    \label{figure_1}
\end{figure*}

\section{Introduction (Problem Statement)}

Large Language Models (LLMs) have rapidly advanced in recent years, demonstrating impressive capabilities across a range of tasks—from natural language understanding and translation to code generation and complex reasoning~\cite{brown2020language, chowdhery2022palm, openai2023gpt4, bai2023reasoning, chen2021evaluating}. With architectural improvements and techniques like instruction tuning, chain-of-thought (CoT) prompting~\cite{wei2022chain}, and reinforcement learning from human feedback, modern LLMs have achieved state-of-the-art performance on benchmarks once thought to require human-level intelligence.

Yet, despite their impressive breadth, LLMs remain vulnerable to adversarial inputs. Small changes in phrasing—often semantically irrelevant—can dramatically alter model behavior, especially in sensitive applications such as code generation. This vulnerability is not merely cosmetic: subtle prompt perturbations can cause models to ignore logic, omit constraints, or even produce insecure code. As prior work on DeceptPrompt \cite{wu2023deceptprompt} and prompt injection attacks has shown, these models often rely on superficial patterns in prompts, rendering them brittle in the face of real-world input variability.

Recent efforts have emphasized the emergent reasoning capabilities of LLMs. Through techniques like CoT prompting, models can now tackle tasks involving multi-step computation, logical inference, and algorithmic synthesis. These developments have sparked hope that LLMs may be evolving from stochastic parrots into true reasoners. However, the key question remains: do LLMs genuinely reason—or do they simulate reasoning only when cues align with familiar templates? This distinction is crucial, especially when evaluating robustness under linguistic or contextual variation.

To rigorously study this question, Mirzadeh et al. (2024) proposed GSM-Symbolic (\cite{mirzadeh2024gsm}), a benchmark that perturbs math word problems using symbolic and semantic transformations while keeping their logical structure intact. Their results revealed significant accuracy drops under benign modifications—suggesting that reasoning performance is highly sensitive to surface form, even when task semantics remain unchanged.

In this work, we extend this line of inquiry to the domain of code generation. We introduce \sys(CoCC), a systematic framework for evaluating reasoning robustness in LLMs using semantically aligned yet adversarial prompt perturbations. Our perturbations span a diverse set of transformations—including storytelling, gamification, domain shifts, distracting constraints, and negation—that vary in narrative structure, lexical framing, and semantic drift. Crucially, each transformation preserves the core logic of the original task, allowing us to isolate how surface-level changes impact reasoning fidelity.

As shown in Figure~\ref{figure_1}, our methodology reveals striking differences in LLM behavior under these controlled perturbations. Some models collapse under low-preservation rewrites, while others paradoxically improve with narrative scaffolding. Unlike prior benchmarks focused on clean accuracy, our framework probes the cognitive stability of LLMs—highlighting the need for robustness evaluations that account for linguistic diversity, adversarial intent, and human-aligned prompting.

In the paper, we successfully implemented six distinct perturbation methods—numeric perturbation, semantic clause injection, storytelling, gamification, domain shift, and negation. Among existing benchmarks on code generation tasks~\cite{chen2021evaluating, austin2021program}, we applied them to a curated set of 100 LeetCode-style problems from the LiveCodeBench dataset~\cite{jain2024livecodebench}, resulting in 700 perturbed instances. We evaluated nine state-of-the-art LLMs using Pass@1~\cite{jain2024livecodebench} (adapted from LiveCodeBench), code correctness via compilation-based execution, and manual inspection. Our results showed that performance on clean prompts ranged from 95.0\% (Gemini-2.5-Flash) ~\cite{google2024gemini} to 17.0\% (DeepSeek-Coder-1.3B) ~\cite{guo2025deepseek}, while some perturbations like storytelling improved accuracy for specific models (e.g., +23.5\% for LLaMA-3.1-Instruct ~\cite{meta2024llama3}). In contrast, low-preservation attacks such as negation degraded performance by over 50\% in several cases. 
We conducted a detailed ablation study and introduced semantic perturbation robustness as a new evaluation axis to characterize model sensitivity to benign linguistic changes.

The relevance of this work is twofold. First, as LLMs become integrated into development workflows and coding assistants, ensuring their reliability under diverse and imperfect inputs becomes paramount. Second, for the broader NLP community, our findings offer a new lens through which to assess model generalization—not just in terms of accuracy, but in terms of behavioral consistency under semantic perturbation. We believe this framework can serve as a foundation for future research on trustworthy and interpretable LLM reasoning.

%% file: sections/related_works.tex
\section{Related work}

The challenge of evaluating and ensuring true reasoning capabilities in Large Language Models (LLMs), particularly for complex tasks such as code generation, has been approached from several angles. While LLMs have demonstrated strong performance across multiple NLP tasks, significant challenges remain in ensuring their robustness, reliability, and the actual depth of their reasoning quality. Our work, builds upon and extends prior research in three key areas that inform our study on reasoning failures under adversarial prompt perturbations: the nature of reasoning in LLMs, specific advancements and evaluations in LLM-based code generation, and existing research on attacks against LLM reasoning.

\subsection{Reasoning Large Language Models}

Reasoning in LLMs encompasses analytical tasks like math problem-solving, code generation, and logical inference. While traditional models often rely on pattern recognition, reasoning-enhanced models increasingly use techniques such as reasoning tokens and Chain-of-Thought (CoT) prompting~\cite{wei2022chain}. Reasoning tokens are designed as explicit markers within the training process to guide LLMs toward more structured reasoning, an approach that has shown improvements in handling complex reasoning challenges, especially in arithmetic and code-related tasks. CoT prompting~\cite{wei2022chain} and its derivatives, like Zero-shot-CoT~\cite{wang2023plan} and Program of Thoughts~\cite{chen2022program} (which aims to separate computation from reasoning), have demonstrably enhanced multi-step reasoning. However, while CoT can improve performance on standard benchmarks, our work with CoCC investigates a critical subsequent question: whether such induced reasoning chains remain stable and effective when the problem's surface structure is perturbed, even if the core logic is preserved. This inquiry is crucial for understanding the true depth and transferability of the elicited reasoning.

Recent surveys and comprehensive reviews~\cite{mondorf2404beyond, prasad2023receval, golovneva2022roscoe} have further explored the reasoning behavior of LLMs beyond simple accuracy metrics, highlighting a prevalent concern: LLMs often rely on surface-level correlations rather than genuine logical inference. For instance, models might associate certain keywords or phrasing patterns with specific solution templates. CoCC directly tests this dependency by altering these patterns through methods like Storytelling, Gamification, and Domain Shift, all while maintaining the underlying semantic integrity of the task. This reliance on potentially superficial cues, even with advanced prompting techniques, underscores the need for evaluations like ours. In this work, we therefore aim to evaluate reasoning through more nuanced behavioral analyses rather than just final answer correctness, focusing on behavioral consistency under controlled linguistic and structural variations.

\subsection{LLM for Code Generation}

LLMs have shown remarkable capabilities in code generation, assisting developers by translating natural language descriptions into executable code. A comprehensive survey by Jiang et al.~\cite{jiang2024llmcodegen} categorizes recent advancements in Code LLMs, covering aspects such as dataset curation, benchmark evaluations, and ethical implications. Evaluating the performance of LLMs in code generation remains a critical research area. Traditional methods primarily rely on execution-based metrics such as pass@k, which evaluates how often the generated code successfully runs and passes test cases. More recent LLM-based evaluation frameworks, such as CODEJUDGE~\cite{tong2024codejudge}, leverage language models themselves to assess the semantic correctness of generated code.

While these methods are valuable, metrics like pass@k primarily assess functional correctness on standard problem formulations, and LLM-based evaluators like CODEJUDGE, though assessing semantic correctness, may not fully capture how models respond when the presentation of the problem itself is adversarially yet semantically perturbed. Our study aligns with the goal of robust evaluation but extends it by focusing on the stability of the reasoning process leading to code generation when faced with such input variations. CoCC specifically assesses the impact of these adversarial modifications on the reasoning tasks inherent in code generation, probing whether the logical process itself is resilient.

\subsection{Reasoning LLM Attacks}

The robustness of LLMs has been scrutinized through various adversarial attack methodologies. Previous research has explored how seemingly small perturbations, such as typographical errors~\cite{gan2024reasoning} or certain types of adversarial prompts~\cite{wu2023deceptprompt}, can significantly degrade model performance. The GSM-Symbolic benchmark~\cite{mirzadeh2024gsm}, for instance, demonstrated that slight modifications to mathematical problem structures, such as altering numerical values or adding irrelevant clauses, can lead to drastic drops in accuracy, suggesting a fundamental brittleness in LLM reasoning when faced with inputs that deviate from learned patterns.

Adversarial attacks on code generation models have also been investigated. The DeceptPrompt framework~\cite{wu2023deceptprompt} reveals that benign-looking natural language modifications can induce LLMs to generate insecure or incorrect code, exposing vulnerabilities in real-world applications. While DeceptPrompt reveals vulnerabilities leading to insecure code, CoCC explores a complementary set of semantically-focused perturbations (e.g., Storytelling, Gamification, Domain Shift, Distracting Constraints). These are designed not primarily to induce insecurity, but to test the stability and fidelity of the underlying reasoning chain required for correct code generation when familiar problem cues are altered. Our study examines how these adversarial modifications—ranging from narrative restructuring to the injection of logically inert information—impact reasoning models. A key distinction in our methodology is the emphasis on perturbations that, while adversarial in their structure or framing, are designed to be semantically faithful to the original problem's core logic. This allows us to isolate reasoning failures that are not due to misunderstanding a completely different task, but rather to an inability to reason robustly when the problem presentation changes.

In summary, prior work has established the potential of LLMs in complex reasoning and code generation, yet has also consistently highlighted their brittleness and potential over-reliance on learned superficial patterns. Existing evaluation methods and attack strategies provide valuable insights into model capabilities and vulnerabilities. However, a gap remains in systematically understanding how LLMs' reasoning chains for code generation withstand diverse, semantically-grounded perturbations that alter problem presentation without changing the core logic. CoCC addresses this by introducing a novel framework and a suite of perturbation techniques specifically designed to probe these nuanced aspects of reasoning robustness in the domain of code generation.
% By extending prior work on LLM robustness, we aim to provide a more comprehensive understanding of how these models respond to adversarially crafted inputs.\\

% Overall, our research aims to contribute to the growing discourse on LLM reliability by systematically evaluating their reasoning under adversarial conditions. By leveraging LLM-based evaluation methodologies and benchmarking our findings against prior studies, we plan to advance the development of more resilient reasoning models.

%% file: sections/approach.tex
\section{Methodology}
\label{sec:method}

To systematically evaluate how semantically controlled modifications affect the code generation abilities of LLMs, we design a framework that perturbs existing coding problems through a set of natural language transformations. Our method centers on applying meaning-preserving or minimally diverging rewrites to problems, without altering their underlying logic or difficulty.

\subsection{Problem Perturbation Framework}
Unlike prior works~\cite{mu2025evaluate, zhu2025uncertainty} that focus solely on adversarial attacks or CoT reasoning, our framework introduces \textbf{semantically aligned rewrites} of coding problems that preserve the structure of the original coding problem while varying its linguistic, narrative, or logical framing. Inspired by recent studies on symbolic noise~\citep{mirzadeh2024gsm}, narrative-based jailbreaks~\citep{shen2024voice, song2025dagger}, and deceptive prompt manipulations~\citep{wu2023deceptprompt}, we propose four transformation strategies (or "attacks") that vary in their degree of logical preservation --- ranging from fully equivalent reformulations (e.g., storytelling) to objective inversions (e.g., negation). Each strategy is instantiated via a consistent prompting template passed to LLaMA 3.1 8B-Instruct that guides the rewriting of the original problem, while explicitly preserving key components: Input, Output, Explanation, Examples, and Constraints. We designed different types of tests to check the language model's reasoning abilities in several key areas. Storytelling and Gamification test the model's ability to extract formal logic from informal, human-centric narratives, a common scenario in real-world user prompts. Domain Shift evaluates the core generalization capability by testing if the model's algorithmic knowledge is tied to specific terminological contexts. Distracting Constraints and Example Perturbation directly challenge the model's attentional mechanisms and its reliance on superficial pattern-matching versus robust logical deduction. Finally, the Negation transformations serve as a stress test for true logical manipulation, assessing whether models can invert a learned reasoning process rather than merely retrieving a familiar one. Together, these transformations cover a spectrum from plausible linguistic variation to direct logical challenge.

\subsection{Perturbation Methods}
In this subsection, we thoroughly describe each of the problem perturbation methods.

\input{tab/tab_preservation_score}

\begin{figure*}[ht!]
    \centering
    \begin{tcolorbox}[
        enhanced,
        colframe=blue!50!black,
        colback=blue!5,
        coltitle=white,
        colbacktitle=blue!50!black,
        width=\textwidth,
        arc=3mm,
        boxrule=0.8mm,
        drop shadow,
        title=\small\bfseries Storytelling Prompt Template,
        fonttitle=\bfseries\small
    ]
    {\small
\textbf{Instruction:} Rewrite the problem in a storytelling format, while preserving its logic.\\[0.5em]
\textbf{Guidelines:}
\begin{itemize}
    \item Keep \textbf{Input}, \textbf{Output}, \textbf{Explanation}, \textbf{Example}, and \textbf{Constraints} sections unchanged.
    \item Frame the problem as an engaging story or programming narrative.
\end{itemize}
\texttt{\{original\_content\}}\\
\textit{No additional explanation needed. Just output the modified problem.}
    }
    \end{tcolorbox}
    \caption{\small Prompt template for Storytelling transformation.}
    \label{fig_storytelling_prompt}
\end{figure*}

\paragraph{Storytelling.} Motivated by narrative jailbreaks in safety research~\citep{shen2024voice, song2025dagger, chan2025speak}, this method reframes the problem as a story or adventure. The prompt preserves all critical components but introduces a narrative wrapper that may change surface-level structure. We use this transformation to study whether storytelling facilitates or hinders LLM reasoning in algorithmic contexts. The \textit{Storytelling} prompt template explicitly instructs the models to "make the introduction engaging and fun", but \textbf{not to alter} the technical sections such as examples, constraints, or explanation by explicitly mentioning \textit{"keep the original intent and structure"} in the prompt. This transformation tests whether LLMs benefit from more natural, human-readable phrasings while still preserving problem semantics (see Figure~\ref{fig_storytelling_prompt}).

\paragraph{Gamification.} This transformation wraps the task in a challenge-based format (e.g., player or robots solving the task) while keeping the logic intact. Similar to storytelling, it introduces anthropomorphic or thematic context, drawing inspiration from instruction-tuning setups that present tasks in game-like scenarios~\citep{wu2023deceptprompt}. We evaluate whether such gamified inputs boosts engagement or distracts models from formal reasoning (see Figure~\ref{fig:gamification_prompt}).

\paragraph{Distracting Constraints.} Inspired by the symbolic injection strategies in GSM-Symbolic~\citep{mirzadeh2024gsm}, this method appends irrelevant but plausible-sounding constraints (e.g., \textit{"input is a palindrome"}) to the problem. While logically inert, these constraints test the model's ability to filter out irrelevant linguistic noise. Unlike classic adversarial examples, these do not change task correctness but increase input ambiguity (see Figure~\ref{fig:distracting_constraints_prompt}).

\paragraph{Domain Shift.} This strategy replaces tokens and terminology from one context to another (e.g., "\textit{array of integers}: becomes \textit{"daily revenue"}) or replacing terms with domain-specific equivalents (e.g., convert numbers to elevator floors, jobs, floors, days) (see Figure~\ref{fig:domain_shift_prompt}).

\subsection{Logical Preservation Scoring}
To quantify the degree of semantic deviation induced by each transformation, we developed a \textbf{Logical Preservation Score}. We employed the Claude-3.7-Sonnet model as an automated evaluator to ensure consistent and scalable scoring across all 700 perturbed instances. This approach mitigates potential human rater fatigue and subjectivity. The scoring is guided by a detailed rubric-based prompt (see Appendix~\ref{appendix_preservation}, Figure~\ref{fig:logical_preservation_prompt}) that instructs the model to rate preservation on a scale of 
\textbf{1-10}, where 10 indicates perfect logical alignment and 1 indicates a complete semantic shift. These scores are used in later analysis to understand how semantic drift correlates with performance degradation.

%% file: tab/tab_preservation_score.tex
\begin{table}[ht]
\centering
\small
\begin{tabular}{lr}
\toprule
\textbf{Attack Type} & \textbf{Score} \\
\midrule
Storytelling & 8.89 \\
Gamification & 8.01 \\
Domain Shift & 7.07 \\
Example Perturbation & 6.71 \\
Distracting Constraints & 3.82 \\
% Negation Objective (Soft) & 2.56 \\
% Negation Objective & 1.83 \\
\bottomrule
\end{tabular}
\caption{Preservation Scores for Different Attack Types}
\label{tab:preservation_score}
\end{table}

%% file: sections/data.tex
\section{Experimental Setting}
% \begin{figure*}[ht]
%     \centering
%     \includegraphics[width=\textwidth]{figs/fig: plot_1.png} % Adjust width as needed
%     \caption{Accuracy comparison across different models and attack types for Easy, Medium, and Hard LeetCode questions.}
%     \label{fig:example} % Reference in text: \ref{fig:example}
% \end{figure*}

\input{tab/tab_clean_acc}
\input{tab/tab_attack_acc}
\input{tab/tab_attack_acc_split_difficulty}

\subsection{Dataset and Evaluation Metrics}
We conduct our experiments using the LiveCodeBench dataset~\cite{jain2024livecodebench}, a recent benchmark designed to evaluate LLM performance on code generation under interactive and test-driven settings. For our study, we select a curated subset of 100 LeetCode problems from the \textbf{Code Generation} split of the LiveCodeBench, which provides rich problem descriptions and test cases suitable for both modifications and evaluation. To assess model performance, we adopt the standard \textbf{Pass@1} metric~\cite{jain2024livecodebench}, which measures the proportion of problems for which the model's first generated solution passes all provided test cases.

\subsection{Model Selection}
We evaluate both \textbf{black-box API models} and \textbf{open-source models} spanning general-purpose reasoning and code-specialized capabilities. For consistency with prior work, we include models from the \textbf{LiveCodeBench leaderboard}\footnote{\url{https://livecodebench.github.io/leaderboard.html}} as well as strong code-focused baselines not originally part of the benchmark.

We test four proprietary black-box LLMs known for their reasoning abilities: \textbf{Gemini-2.5-Flash-Preview},  \textbf{Gemini-2.0-Flash}~\cite{google2024gemini}, \textbf{Claude-3.7-Sonnet}, and \textbf{Claude-3-Haiku}~\cite{anthropic2024claude35sonnet}. To assess open-source reasoning, we include \textbf{DeepSeek-R1-Distill-Qwen-7B} and \textbf{-14B}~\cite{guo2025deepseek}, which are distilled DeepSeek-R1 models with robust reasoning capabilities and native support for \textit{reasoning token generation}, facilitating fine-grained trace analysis. For code-focused models, we evaluate: \textbf{Qwen2.5-Coder} and \textbf{DeepSeek-Coder-33B}, both of which are optimized for code generation but not explicitly tuned for reasoning. Finally, we include the \textbf{LLaMA-3.1-8B-Instruct}~\cite{meta2024llama3}, a general-purpose instruction-tuned model. Although not code- nor reasoning-specialized, LLaMA-3.1 performs competitively and supports inference-time CoT prompt engineering, making it a valuable reference point in our analysis.

%% file: tab/tab_clean_acc.tex
\begin{table}[ht]
\centering
\resizebox{\columnwidth}{!}{%
\begin{tabular}{l|rrr|r}
\toprule
\textbf{Model} & \textbf{Easy} & \textbf{Medium} & \textbf{Hard} & \textbf{Overall} \\
\midrule
Gemini-2.5-Flash & 100.0 & 98.0 & 76.5 & 95.0 \\
Claude-3.7-Sonnet & 100.0 & 92.0 & 64.7 & 90.0 \\
DeepSeek-14B & 97.0 & 92.0 & 58.8 & 88.0 \\
Gemini-2.0-Flash & 97.0 & 86.0 & 47.1 & 83.0 \\
DeepSeek-7B & 90.9 & 70.0 & 35.3 & 71.0 \\
Qwen2.5-Coder & 81.8 & 52.0 & 35.3 & 59.0 \\
DeepSeek-Coder-33B & 84.8 & 48.0 & 11.8 & 54.0 \\
Claude-3-Haiku & 51.5 & 32.0 & 41.2 & 40.0 \\
LLaMA-3.1-8B-Instruct & 36.4 & 12.0 & 5.9 & 19.0 \\
DeepSeek-Coder-1.3B & 36.4 & 8.0 & 5.9 & 17.0 \\
\midrule
\textbf{Avg. Accuracy} & 77.6 & 59.9 & 38.3 & 67.6 \\
\bottomrule
\end{tabular}
}
\caption{Accuracy (\%) of models across different difficulty levels under the no attack setting.}
\label{tab:accuracy_clean}
\end{table}

%% file: tab/tab_attack_acc.tex
\begin{table*}[ht]
\centering
\resizebox{\textwidth}{!}{%
\begin{tabular}{l|rrrrr|r}
\toprule
\textbf{Models} & \makecell[r]{\textbf{Distracting}\\\textbf{Constraints}} & \makecell[r]{\textbf{Domain}\\\textbf{Shift}} & \makecell[r]{\textbf{Example}\\\textbf{Perturbation}} & \textbf{Gamification} & \textbf{Storytelling} & \textbf{Avg. Acc.} \\
\midrule
Gemini 2.5 Flash & 95.48\% \textcolor{teal}{(+0.5\%)} & 89.81\% \textcolor{purple}{(-5.2\%)} & 93.09\% \textcolor{purple}{(-1.9\%)} & 96.93\% \textcolor{teal}{(+1.9\%)} & 97.37\% \textcolor{teal}{(+2.4\%)} & 94.94\% \\
Gemini 2.0 Flash & 88.78\% \textcolor{teal}{(+5.8\%)} & 72.47\% \textcolor{purple}{(-10.5\%)} & 95.02\% \textcolor{teal}{(+12.0\%)} & 89.76\% \textcolor{teal}{(+6.8\%)} & 90.38\% \textcolor{teal}{(+7.4\%)} & 87.28\% \\
DeepSeek-14B & 83.84\% \textcolor{purple}{(-4.2\%)} & 75.53\% \textcolor{purple}{(-12.5\%)} & 84.10\% \textcolor{purple}{(-3.9\%)} & 86.76\% \textcolor{purple}{(-1.2\%)} & 91.16\% \textcolor{teal}{(+3.2\%)} & 84.28\% \\
Qwen2.5-Coder & 65.03\% \textcolor{teal}{(+6.0\%)} & 68.33\% \textcolor{teal}{(+9.3\%)} & 83.51\% \textcolor{teal}{(+24.5\%)} & 70.11\% \textcolor{teal}{(+11.1\%)} & 79.69\% \textcolor{teal}{(+20.7\%)} & 73.33\% \\
DeepSeek-7B & 65.06\% \textcolor{purple}{(-5.9\%)} & 58.28\% \textcolor{purple}{(-12.7\%)} & 71.18\% \textcolor{teal}{(+0.2\%)} & 73.30\% \textcolor{teal}{(+2.3\%)} & 82.99\% \textcolor{teal}{(+12.0\%)} & 70.16\% \\
DeepSeek-Coder-33B & 58.44\% \textcolor{teal}{(+4.4\%)} & 55.26\% \textcolor{teal}{(+1.3\%)} & 70.00\% \textcolor{teal}{(+16.0\%)} & 67.44\% \textcolor{teal}{(+13.4\%)} & 70.45\% \textcolor{teal}{(+16.5\%)} & 64.32\% \\
Claude-3.7-Sonnet & 47.89\% \textcolor{purple}{(-42.1\%)} & 35.66\% \textcolor{purple}{(-54.3\%)} & 62.87\% \textcolor{purple}{(-27.1\%)} & 50.00\% \textcolor{purple}{(-40.0\%)} & 63.35\% \textcolor{purple}{(-26.6\%)} & 51.95\% \\
Claude-3-Haiku & 41.04\% \textcolor{teal}{(+1.0\%)} & 37.98\% \textcolor{purple}{(-2.0\%)} & 60.12\% \textcolor{teal}{(+20.1\%)} & 45.14\% \textcolor{teal}{(+5.1\%)} & 57.69\% \textcolor{teal}{(+17.7\%)} & 48.39\% \\
LLaMA3-8B-Instruct & 37.59\% \textcolor{teal}{(+18.6\%)} & 22.03\% \textcolor{teal}{(+3.0\%)} & 54.30\% \textcolor{teal}{(+35.3\%)} & 37.78\% \textcolor{teal}{(+18.8\%)} & 44.68\% \textcolor{teal}{(+25.7\%)} & 39.28\% \\
\midrule
\textbf{Avg. Acc.} & 64.57\% & 57.70\% & 75.24\% & 68.80\% & 75.64\% & -- \\
\bottomrule
\end{tabular}
}
\caption{Attack accuracy (\%) and delta $\Delta$ from clean performance. Rows are sorted by average accuracy.}
\label{tab:attack_table_transposed_sorted}
\end{table*}

%% file: tab/tab_attack_acc_split_difficulty.tex
\begin{table*}[ht]
\centering
\small
\begin{tabular}{llrrr}
\toprule
\textbf{Attack} & \textbf{Model} & \textbf{Easy Acc.} & \textbf{Medium Acc.} & \textbf{Hard Acc.} \\
\midrule
\multirow{6}{*}{Storytelling}
 & Gemini-2.5-Flash & 93.9\% \textcolor{purple}{(-6.1\%)} & 96.0\% \textcolor{purple}{(-2.0\%)} & 88.2\% \textcolor{teal}{(+11.7\%)} \\
 & Gemini-2.0-Flash & 90.9\% \textcolor{purple}{(-6.1\%)} & 82.0\% \textcolor{purple}{(-4.0\%)} & 52.9\% \textcolor{teal}{(+5.8\%)} \\
 & Claude-3.7-Sonnet & 48.5\% \textcolor{purple}{(-51.5\%)} & 34.0\% \textcolor{purple}{(-58.0\%)} & 47.1\% \textcolor{purple}{(-17.6\%)} \\
 & Claude-3-Haiku & 45.5\% \textcolor{purple}{(-6.0\%)} & 26.0\% \textcolor{purple}{(-6.0\%)} & 35.3\% \textcolor{purple}{(-5.9\%)} \\
 & DeepSeek-14B & 93.9\% \textcolor{purple}{(-3.1\%)} & 82.0\% \textcolor{purple}{(-10.0\%)} & 52.9\% \textcolor{purple}{(-5.9\%)} \\
 & DeepSeek-7B & 97.0\% \textcolor{teal}{(+6.1\%)} & 62.0\% \textcolor{purple}{(-8.0\%)} & 23.5\% \textcolor{purple}{(-11.8\%)} \\
\midrule

\multirow{6}{*}{Domain Shift}
 & Gemini-2.5-Flash & 89.8\% \textcolor{purple}{(-10.2\%)} & 96.0\% \textcolor{gray}{(+0.0\%)} & 82.4\% \textcolor{teal}{(+5.9\%)} \\
 & Gemini-2.0-Flash & 72.5\% \textcolor{purple}{(-24.5\%)} & 80.0\% \textcolor{purple}{(-6.0\%)} & 58.8\% \textcolor{teal}{(+11.7\%)} \\
 & Claude-3.7-Sonnet & 35.7\% \textcolor{purple}{(-64.3\%)} & 24.0\% \textcolor{purple}{(-68.0\%)} & 11.8\% \textcolor{purple}{(-52.9\%)} \\
 & Claude-3-Haiku & 30.3\% \textcolor{purple}{(-21.2\%)} & 18.0\% \textcolor{purple}{(-14.0\%)} & 5.9\% \textcolor{purple}{(-35.3\%)} \\
 & DeepSeek-14B & 54.5\% \textcolor{purple}{(-42.5\%)} & 62.0\% \textcolor{purple}{(-30.0\%)} & 29.4\% \textcolor{purple}{(-29.4\%)} \\
 & DeepSeek-7B & 45.5\% \textcolor{purple}{(-45.4\%)} & 38.0\% \textcolor{purple}{(-32.0\%)} & 17.6\% \textcolor{purple}{(-17.7\%)} \\
\midrule

\multirow{6}{*}{Example Perturbation}
 & Gemini-2.0-Flash & 97.0\% \textcolor{gray}{(+0.0\%)} & 92.0\% \textcolor{teal}{(+6.0\%)} & 64.7\% \textcolor{teal}{(+17.6\%)} \\
 & Gemini-2.5-Flash & 93.9\% \textcolor{purple}{(-6.1\%)} & 84.0\% \textcolor{purple}{(-14.0\%)} & 70.6\% \textcolor{purple}{(-5.9\%)} \\
 & Claude-3.7-Sonnet & 51.5\% \textcolor{purple}{(-48.5\%)} & 34.0\% \textcolor{purple}{(-58.0\%)} & 23.5\% \textcolor{purple}{(-41.2\%)} \\
 & Claude-3-Haiku & 45.5\% \textcolor{purple}{(-6.0\%)} & 28.0\% \textcolor{purple}{(-4.0\%)} & 35.3\% \textcolor{purple}{(-5.9\%)} \\
 & DeepSeek-14B & 81.8\% \textcolor{purple}{(-15.2\%)} & 68.0\% \textcolor{purple}{(-24.0\%)} & 47.1\% \textcolor{purple}{(-11.7\%)} \\
 & DeepSeek-7B & 72.7\% \textcolor{purple}{(-18.2\%)} & 42.0\% \textcolor{purple}{(-28.0\%)} & 35.3\% \textcolor{gray}{(+0.0\%)} \\
\midrule
\multirow{6}{*}{Distracting Constraints}
 & Gemini-2.5-Flash & 84.8\% \textcolor{purple}{(-15.2\%)} & 96.0\% \textcolor{purple}{(-2.0\%)} & 82.4\% \textcolor{teal}{(+5.9\%)} \\
 & Gemini-2.0-Flash & 81.8\% \textcolor{purple}{(-15.2\%)} & 80.0\% \textcolor{purple}{(-6.0\%)} & 58.8\% \textcolor{teal}{(+11.7\%)} \\
 & Claude-3.7-Sonnet & 36.4\% \textcolor{purple}{(-63.6\%)} & 24.0\% \textcolor{purple}{(-68.0\%)} & 11.8\% \textcolor{purple}{(-52.9\%)} \\
 & Claude-3-Haiku & 36.4\% \textcolor{purple}{(-15.1\%)} & 12.0\% \textcolor{purple}{(-20.0\%)} & 17.6\% \textcolor{purple}{(-23.6\%)} \\
 & DeepSeek-14B & 78.8\% \textcolor{purple}{(-18.2\%)} & 70.0\% \textcolor{purple}{(-22.0\%)} & 41.2\% \textcolor{purple}{(-17.6\%)} \\
 & DeepSeek-7B & 57.6\% \textcolor{purple}{(-33.3\%)} & 36.0\% \textcolor{purple}{(-34.0\%)} & 29.4\% \textcolor{purple}{(-5.9\%)} \\
\midrule
\multirow{6}{*}{Gamification}
 & Gemini-2.5-Flash & 100.0\% \textcolor{gray}{(+0.0\%)} & 92.0\% \textcolor{purple}{(-6.0\%)} & 88.2\% \textcolor{teal}{(+11.7\%)} \\
 & Gemini-2.0-Flash & 97.0\% \textcolor{gray}{(+0.0\%)} & 74.0\% \textcolor{purple}{(-12.0\%)} & 58.8\% \textcolor{teal}{(+11.7\%)} \\
 & Claude-3.7-Sonnet & 48.5\% \textcolor{purple}{(-51.5\%)} & 30.0\% \textcolor{purple}{(-62.0\%)} & 23.5\% \textcolor{purple}{(-41.2\%)} \\
 & Claude-3-Haiku & 51.5\% \textcolor{gray}{(+0.0\%)} & 34.0\% \textcolor{teal}{(+2.0\%)} & 35.3\% \textcolor{purple}{(-5.9\%)} \\
 & DeepSeek-14B & 84.8\% \textcolor{purple}{(-12.2\%)} & 64.0\% \textcolor{purple}{(-28.0\%)} & 52.9\% \textcolor{purple}{(-5.9\%)} \\
 & DeepSeek-7B & 75.8\% \textcolor{purple}{(-15.1\%)} & 46.0\% \textcolor{purple}{(-24.0\%)} & 29.4\% \textcolor{purple}{(-5.9\%)} \\
\bottomrule
\end{tabular}
\caption{Accuracy (\%) for \textbf{Reasoning-focused LLMs} under various attacks, broken down by difficulty level.}
\label{tab:reasoning_models_difficulty}
\end{table*}

\begin{table*}[ht]
\centering
\small
\begin{tabular}{llr|r|r}
\toprule
\textbf{Attack} & \textbf{Model} & \textbf{Easy Acc.} & \textbf{Medium Acc.} & \textbf{Hard Acc.} \\
\midrule
\multirow{4}{*}{Storytelling}
 & Qwen2.5-Coder & 84.8\% \textcolor{teal}{(+3.0\%)} & 54.0\% \textcolor{teal}{(+2.0\%)} & 35.3\% \textcolor{gray}{(+0.0\%)} \\
 & DeepSeek-Coder-33B & 81.8\% \textcolor{purple}{(-3.0\%)} & 34.0\% \textcolor{purple}{(-14.0\%)} & 23.5\% \textcolor{teal}{(+11.7\%)} \\
 & LLaMA3-8B-Instruct & 45.5\% \textcolor{teal}{(+9.1\%)} & 4.0\% \textcolor{purple}{(-8.0\%)} & 29.4\% \textcolor{teal}{(+23.5\%)} \\
\midrule
\multirow{4}{*}{Example Perturbation}
 & Qwen2.5-Coder & 84.8\% \textcolor{teal}{(+3.0\%)} & 66.0\% \textcolor{teal}{(+14.0\%)} & 41.2\% \textcolor{teal}{(+5.9\%)} \\
 & LLaMA3-8B-Instruct & 45.5\% \textcolor{teal}{(+9.1\%)} & 28.0\% \textcolor{teal}{(+16.0\%)} & 11.8\% \textcolor{teal}{(+5.9\%)} \\
 & DeepSeek-Coder-33B & 72.7\% \textcolor{purple}{(-12.1\%)} & 44.0\% \textcolor{purple}{(-4.0\%)} & 17.6\% \textcolor{teal}{(+5.8\%)} \\
\midrule
\multirow{4}{*}{Distracting Constraints}
 & Qwen2.5-Coder & 57.6\% \textcolor{purple}{(-24.2\%)} & 42.0\% \textcolor{purple}{(-10.0\%)} & 17.6\% \textcolor{purple}{(-17.7\%)} \\
 & DeepSeek-Coder-33B & 54.5\% \textcolor{purple}{(-30.3\%)} & 26.0\% \textcolor{purple}{(-22.0\%)} & 29.4\% \textcolor{teal}{(+17.6\%)} \\
 & LLaMA3-8B-Instruct & 30.3\% \textcolor{purple}{(-6.1\%)} & 8.0\% \textcolor{purple}{(-4.0\%)} & 17.6\% \textcolor{teal}{(+11.7\%)} \\
\midrule
\multirow{4}{*}{Gamification}
 & Qwen2.5-Coder & 75.8\% \textcolor{purple}{(-6.0\%)} & 52.0\% \textcolor{gray}{(+0.0\%)} & 23.5\% \textcolor{purple}{(-11.8\%)} \\
 & DeepSeek-Coder-33B & 69.7\% \textcolor{purple}{(-15.1\%)} & 50.0\% \textcolor{teal}{(+2.0\%)} & 35.3\% \textcolor{teal}{(+23.5\%)} \\
 & LLaMA3-8B-Instruct & 42.4\% \textcolor{teal}{(+6.0\%)} & 20.0\% \textcolor{teal}{(+8.0\%)} & 23.5\% \textcolor{teal}{(+17.6\%)} \\
\midrule
 \multirow{4}{*}{Domain Shift}
 & Qwen2.5-Coder & 64.3\% \textcolor{purple}{(-17.5\%)} & 41.2\% \textcolor{purple}{(-10.8\%)} & 20.0\% \textcolor{purple}{(-15.3\%)} \\
 & DeepSeek-Coder-33B & 48.5\% \textcolor{purple}{(-36.3\%)} & 26.0\% \textcolor{purple}{(-22.0\%)} & 17.6\% \textcolor{teal}{(+5.8\%)} \\
 & LLaMA3-8B-Instruct & 15.2\% \textcolor{purple}{(-21.2\%)} & 2.0\% \textcolor{purple}{(-10.0\%)} & 11.8\% \textcolor{teal}{(+5.9\%)} \\

\bottomrule
\end{tabular}
\caption{Accuracy (\%) for Code Generation Models under various attacks.}
\label{tab:code_models}
\end{table*}

%% file: sections/experiments.tex
\section{Results and Analysis}
In this section, we present both aggregate and manual case-level results from our evaluation of LLMs under perturbations. While overall accuracy and reasoning scores provide a high-level view of model robustness, a closer examination of specific instances reveals nuanced patterns. We identify cases where perturbations help models reason better and others where they introduce confusion, highlighting the diverse impacts of prompt formulation on code generation.

\subsection{Clean Performance Overview}
We begin by benchmarking all models on the original (unmodified) version of 100 LeetCode-style problems (Table~\ref{tab:accuracy_clean}). As expected, models optimized for instruction-following and reasoning, such as Gemini-2.5-Flash (95.0\% overall) and Claude-3.7-Sonnet (90.0\%), outperform code-specific models like Qwen2.5-Coder (59.0\%) and DeepSeek-Coder-33B (54.0\%). ON average, performance declines sharply from Easy (77.6\%) to Hard (38.3\%) tasks, reflecting the inherent challenge of generalization across reasoning complexity.

\subsection{Perturbation Resilience and Gains}
Table~\ref{tab:attack_table_transposed_sorted} reveals that \textbf{semantic perturbations do not uniformly degrade performance}. On the contrary, several perturbation types --- particularly Storytelling and Gamification --- \textit{consistently improve performance}, especially for models tuned for human-aligned reasoning.

Notably, Gemini-2.0-Flash, which ranks 4th in clean accuracy, \textbf{outperforms its original baseline} on four perturbation types, improving up to +12.0\% on Example Perturbation. Similarly Qwen2.5-Coder and LLaMA-3.1, both non-reasoning-specialized, exhibit large performance gains (+24.5\% and +35.3\%, respectively) under storytelling-style or pattern-breaking rewrites. Based on these findings, we noticed that certain models benefit from more natural or decentered problem formulations. This suggests that rigid, minimalistic problem statements may actually underutlize LLM capabilities. \\
This is further supported by our manual inspection of Gemini-2.5-Flash outputs, where in 11 cases of Distracting Constraints attack, the perturbed versions with added constraints produced correct solutions while the original did not. These constraints, though labeled as "distracting", may have made implicit assumptions explicit, improving clarity. This aligns with findings by \citet{xu2024contextual}, who show that structured context in prompts can enhance generation accuracy.
 This finding challenges prior assumptions that natural-language verbosity of format deviation is always detrimental~\cite{zhu2024promptbench}. Instead, we observe that semantically richer prompts can elicit \textbf{better program synthesis}. 

\subsection{Semantic Robustness} 
Despite general trends, not all models are equally resilient to semantic shift. Claude-3.7-Sonnet exhibits massive accuracy degradation under low-preservation transformations, losing up to \textbf{-68.0\%} in medium difficulty tasks for Distracting Constraints. In contrast, Gemini-2.5-Flash and DeepSeek-14B are remarkably stable, showing $<$5\% deviation across all tasks. Through these experimental results, we demonstrate a novel axis of evaluation: Semantic Perturbation Robustness --- defined as the standard deviation in performance across logic-preserving rewrites. This axis captures a model's ability to maintain performance under benign yet diverse formulations of the same task. We argue that this robustness is as critical as clean accuracy, especially in real-world applications where user queries are linguistically varied and imperfectly phrased.\\
This phenomenon is visible in models like Claude-3.7-Sonnet, where code generated from perturbed prompts is often simpler and less complete compared to the original. It appears that the model relies on surface-level patterns in the original prompt to retrieve structured code, and perturbations disrupt this retrieval behavior. As \citet{cai2024entropy} observe, such perturbations likely increase token-level entropy and variance, making the model less confident and more prone to errors.

\begin{figure}[ht]
    \centering
    \includegraphics[width=0.90\linewidth]{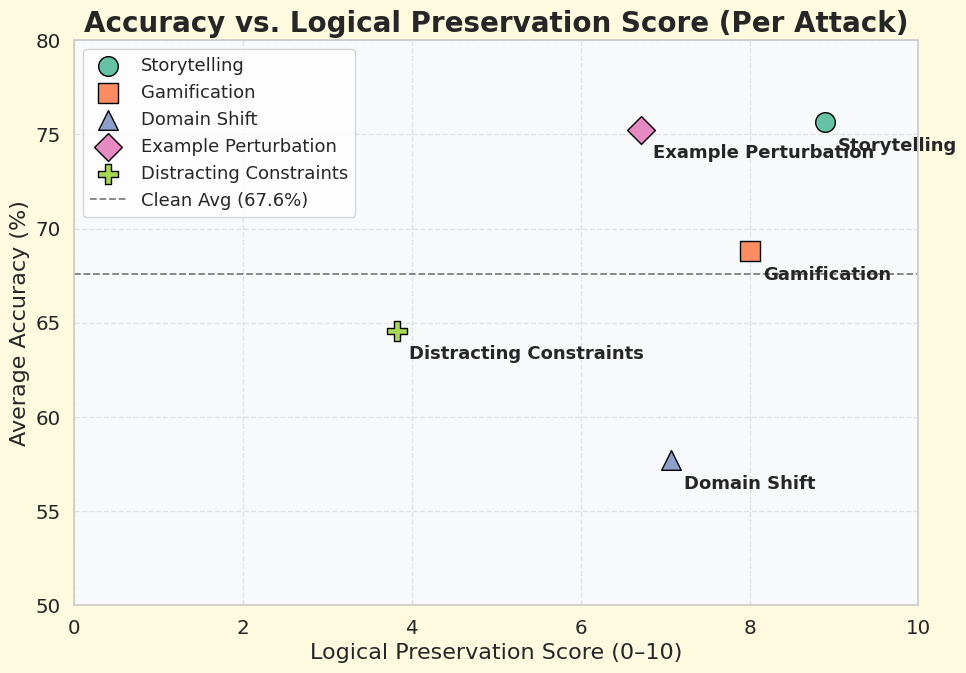}
    \caption{Accuracy vs. Logical Preservation Score for each attack type. Storytelling and Example Perturbation improve performance despite non-trivial rephrasings, while Distracting Constraints degrades accuracy significantly.}
    \label{fig_accuracy_vs_preservation}
\end{figure}

\subsection{Accuracy vs. Logic Preservation}
In Figure~\ref{fig_accuracy_vs_preservation}, we plot average model accuracy against our logical preservation scores (Table~\ref{tab:preservation_score}. A positive, albeit nonlinear, correlation emerges: Storytelling (score: 8.89) and Gamification (8.01) lead to improved or stable performance. Distracting Constraints (3.82) and Negation Objective (1.83) consistently degrade it. However, we noticed that performance does not degrade proportionally with logical drift. Some logically preserved rewrites improve accuracy, while a single line of structurally trivial edits (like Distracting Constraints) dramatically hurt it. This suggests that cognitive alignment --- not logical alignment alone --- governs LLM success. LLMs are more sensitive to semantic distractors than to narrative abstraction.

\subsection{Difficulty Analysis}
As shown in Table~\ref{tab:reasoning_models_difficulty}, models often retain high accuracy on Easy problems across all attacks, but break under Hard tasks even for high-preservation rewrites. This confirms that reasoning complexity, not just input phrasing, lead to model breakdown. However, storytelling notably \textbf{boosts hard-problem performance} for Gemini-2.5 (+11.7\%) and LLaMA3.1 (+23.5\%), indicating that human-friendly reformulations may facilitate deeper reasoning under challenge. 

\subsection{Errors through Prompt Clarity}

To interpret why some perturbations help while others hinder, we draw on recent findings that connect model uncertainty to error rates. \citet{cai2024entropy} observe that errors in LLMs are more likely to occur at tokens with high entropy and variance regions where the model is unsure of the next step. When we perturb prompts, especially with vague or extraneous constraints, the resulting prompt may distribute the model's attention more diffusely, increasing uncertainty and leading to incorrect generations.

This highlights a dual nature of perturbations: when they disrupt expected patterns without adding structure, they degrade performance; but when they elaborate or scaffold problem context, they may improve reasoning as we see above for certain models in Table~\ref{tab:code_models}. Future models may benefit from tuning not only on correct answers, but on uncertainty aware reasoning paths.

\subsection{Comment-Driven Reasoning Behaviors}

\input{tab/tab_comments}
A notable behavioral pattern emerged when analyzing Claude-3.7-Sonnet's performance under the Gamification prompt variant. While Claude-3.7 demonstrates strong performance under clean conditions (90.0\% overall accuracy; Table~\ref{tab:accuracy_clean}), its accuracy drops dramatically by --40.0\% under Gamification (Table~\ref{tab:attack_table_transposed_sorted}). Manual inspection of model outputs (Table  ~\ref{tab:claude_comment_example}) revealed a consistent shift in behavior: Claude’s code responses on unperturbed prompts routinely included inline comments---clarifying control flow logic, edge case handling, or the purpose of specific variables. In contrast, responses under gamified prompt formulations lacked such explanatory commentary.

The absence of natural language comments appeared to correlate with degraded code quality. Logic that was correctly implemented in the clean setting often became brittle or misaligned under perturbation, indicating reduced internal reasoning coherence. This trend is consistent with findings by Dou et al.~\cite{dou2024whats}, who showed that a higher frequency of comments in LLM-generated code is associated with enhanced reasoning capabilities and fewer bugs. Their results suggest that encouraging comment generation during code synthesis can improve reasoning—even without model fine-tuning.

Our results extend this insight by proposing that comment generation serves not only as a reflection of internal reasoning, but also as a mediating factor in model robustness. When Claude-3.7 omits comments under prompt perturbation, its performance deteriorates significantly---despite the underlying task semantics remaining unchanged. This supports the hypothesis that comment-driven reasoning may serve a stabilizing role in preserving performance across diverse linguistic formulations of the same task.

\subsection{Deeper dive in domain shift}
A detailed analysis of the evaluation logs reveals distinct behavioral patterns. To begin with, Llama-3.1's robustness appears to stem from its capability to abstract the underlying logical task from the superficial narrative. Its reasoning often involved re-stating the problem with the new domain terms and then applying the correct, generalized algorithm, effectively handling the semantic shift (see Appendix C.1). Furthermore, the performance gain for Qwen2.5-Coder suggests that shifting to domains more aligned with common programming tasks or utilizing more concrete terminology (e.g., "employee\_salaries" instead of abstract "nums") might have enhanced its understanding or triggered more relevant learned patterns. It consistently adopted new terminologies while correctly implementing the core logic (see Appendix C.1). On the other hand, Claude's substantial accuracy drop highlights a significant fragility when faced with unfamiliar thematic contexts. In several instances, Claude's reasoning (output) would correctly re-state the objective in the new domain, yet its generated solution\_code would be drastically incomplete, address a much simpler version of the problem, or fail to implement the core preserved logic. For example, in the minIncrementOperations task (originally about making an array "beautiful" based on a sliding window condition, shifted to stabilizing task\_priorities based on a similar window), Claude correctly understood and re-stated the complex windowed objective in the new domain terms. However, its subsequent code implemented a much simpler, incorrect element-wise check, completely missing the required windowed logic (see Appendix C.2). This indicates that while it can process surface-level semantic changes, the mapping to the correct complex algorithmic solution becomes unreliable. DeepSeek-14B, while not failing as comprehensively as Claude, still showed difficulties. Its errors often involved a correct general algorithmic idea but flawed execution within the new domain, such as misinterpreting new terms or making errors in adapting conditions, indicating challenges in accurately translating familiar logic to novel semantic contexts (see Appendix C.1).

\section{Ablation Study}
To probe the boundary of semantic robustness, we conduct an ablation study focused on \textbf{Negation Objective (Hard)} and \textbf{Negation Objective (Soft)} transformations. These rewrites are designed to test whether models can adapt to \textbf{inverted or subtly reversed objectives}, while preserving the format and structural elements of the original problem.

\subsection{Methods}

\paragraph{Negation Objective (Hard).} This method fully inverts the core task (e.g., "maximize" $\rightarrow$ "minimize" or "include" $\rightarrow$ "exclude"). A part of the transformation prompt asks the LLM to: \textit{"rewrite it with the opposite objective ... ensure the examples and output are updated accordingly:} (see Figure~\ref{fig:negation_prompt}). While the input, output, explanation and constraints are preserved, the solution logic must change, requiring the model to recompute from a new problem intent. The preservation score was recorded as 1.83 based on Claude-3.7.

\paragraph{Negation Objective (Soft).} This variant applied milder semantic reversals (e.g., asking for "non-decreasing" instead of "increasing") without flipping the task entirely. The instruction explicitly states: \textit{"Do NOT fundamentally change the problem's task ... adjust the examples minimally"} (see Figure~\ref{fig:soft_negation_prompt}). This design tests whether models are robust to subtle linguistic tweaks that alter logic while retaining solvability. The preservation score was recorded as 2.56 based on Claude-3.7, a little higher than the hard negation.

\subsection{Results}
Table~\ref{tab:ablation} summarizes the impact of negation-based perturbations. Hard negation proves highly disruptive, causing average accuracy drops of over 50\% across all difficulty levels. For instance, Gemini-2.5-Flash drops from 95.0\% to 42.6\%, and Claude-3.7 falls from 90.0\% to 15.0\%. Even instruction-following models struggle to adapt, as reasoning reversal—not just language adaptation—is required. In contrast, soft negation shows more variability, with some models like LLaMA-3.1 improving from 19.0\% to 29.4\%. However, others like DeepSeek-14B and Claude-3.7 still suffer large drops, showing that even subtle logical shifts can disrupt reasoning.

\begin{comment}
% Suggested lead-in:
\textbf{Deeper Analysis of Soft Negation Behaviors:}
The "Negation Objective (Soft)" perturbation (Logical Preservation Score: 2.56) revealed nuanced model responses. Llama-3.1-8B-Instruct notably exhibited a +10.4\% accuracy increase, contrasting with significant drops for Claude-3.7-Sonnet (-64.6\%), Gemini-2.5-Flash-Preview (-53.5\%), and DeepSeek-14B (-86.9\%).~\textcolor{red}{This is not needed as we already mentioned the results above.}
\end{comment}

\subsection{Analysis}
\paragraph{Negation Objective (Soft) Analysis.}
Under soft negation, Llama-3.1-8B-Instruct's improved accuracy was sometimes linked to genuine adaptation when test cases were correctly aligned with the modified textual objective (see Appendix A.1). However, a prevalent issue was test case misalignment: Llama, despite correct logical adaptation to modified text, often failed tests still expecting original problem answers (e.g., for \texttt{sumOfSquares}, Appendix~\ref{app:sumofsquares}). Conversely, models like DeepSeek-14B sometimes "passed" by ignoring the modification, as their test cases also remained unchanged. Failure patterns for other models included partial adaptation, where only parts of the negated logic were implemented (e.g., Gemini-2.5-Flash on \texttt{matrixSum}, Appendix~\ref{app:matrixsum}), and archetype override, where models reverted to known problem solutions despite contradictory modified instructions (e.g., Gemini-2.5-Flash on \texttt{maximumOr}, Appendix A.4). DeepSeek-14B, in particular, consistently ignored modifications. While Llama-3.1 demonstrated better semantic fidelity to textual changes, its overall accuracy increase is a nuanced outcome influenced by both genuine adaptation and the complex interplay with test case consistency. True robustness assessment hinges on meticulously aligned problem prompts and test evaluations.

\paragraph{Negation Objective (Hard) Analysis.}

The Negation Objective (Hard) perturbation led to near-total task failure across models, with reported accuracies ($0\%-17\%$) not reflecting genuine capability in handling inverted objectives due to pervasive \textbf{test case misalignment}. Observed "successes" typically arose when models solved the \textit{original} problem against un-updated test cases (e.g., Qwen2.5-Coder on \texttt{count\_seniors}, Appendix~\ref{app:hardneg-qwen-success}; Gemini-2.5-Flash on \texttt{min\_extra\_char}, Appendix~\ref{app:hardneg-gemini-success}). Conversely, models that correctly implemented simple negated logic often failed these misaligned tests (e.g., Claude, DeepSeek, Gemini on \texttt{count\_seniors}, Appendix~\ref{app:hardneg-correct-fail}). No model consistently demonstrated reasoning reversal. Instead, common behaviors included defaulting to original problem archetypes, code instability (DeepSeek-14B on complex negations, Appendix~\ref{app:hardneg-deepseek-runtime}), and failure to implement correctly inverted logic even when acknowledging the textual change (e.g., on \texttt{matrixSum}, Appendix~\ref{app:hardneg-matrixsum-confusion}). Hard negation thus exposes profound limitations in current LLMs' ability to perform genuine goal and logic reversal, emphasizing the need for rigorously aligned evaluations.

In conclusion, negation-based rewrites expose a critical blind spot in LLM reasoning: the inability to reinterpret or invert logical objectives even when all structural information is preserved. In addition, the model collapse under negation objective suggests that task framing overrides instruction tuning, and that models struggle to recompute from inverse logic. Finally, soft negation reveals the fragility of lexical understanding: small semantic perturbations that humans consider minor can lead to cascading failures in reasoning paths.

\section{Limitation}
One limitation of our work is that we evaluated only a subset of the LiveCodeBench LeetCode-style problems --- specifically, 100 our of 235 problems. Additionally, our study did not cover other task types available in LiveCodeBench, such as code execution, test output prediction, and self-repair. While LiveCodeBench includes a diverse set of problems sources from platforms like AtCoder and Codeforces in addition to LeetCode, our evaluation was restricted to the LeetCode subset. This decision was primarily driven by limited computational resources, as inference time increases substantially with model size, particularly for open-source models. To ensure fair comparison across different models, we maintained a consistent evaluation set. 

Despite these constraints, we conducted a large-scle robustness analysis by generating and evaluating 700 perturbed instances (100 clean problems $\times$ 7 perturbation types), allowing us to systematically assess model behavior under diverse logical modifications.

Another limitation is the exclusion of GPT-series models (e.g., GPT-4), which are among the top-performing proprietary models for code generation according to the LiveCodeBench leaderboard. Due to access limitations, we were unable to include them in our evaluation, leaving their robustness under these perturbations as an open question for future work.

\section{Conclusion}

In this work, we presented CoCC, a systematic framework for evaluating reasoning robustness in LLMs under adversarial prompting in code generation tasks. By introducing semantically plausible but logically challenging perturbations such as storytelling, gamification, distracting constraints, and negation objectives, we revealed key vulnerabilities in current state-of-the-art models. Through comprehensive experiments across nine LLMs and 700 perturbed instances, we found that while some perturbations surprisingly improved model performance, others, especially those involving subtle semantic drift, led to significant reasoning failures.

Our analysis introduced a novel evaluation axis, \emph{semantic perturbation robustness}, which quantifies a model’s ability to generalize across logically equivalent but linguistically diverse inputs. These findings underscore that high clean accuracy does not imply robust reasoning and highlight the need for more cognitively aligned model training.

%% file: tab/tab_comments.tex
\begin{table*}[ht]
  \centering
  % Scale the table by 1.1× (≈10pt→11pt) while keeping it within \textwidth
  \begin{adjustbox}{max width=\textwidth,scale=1}
    \begin{tabular}{l|l|c|c}
      \toprule
      \textbf{Original Prompt Output} & \textbf{Perturbed Prompt Output}
        & \textbf{Original Result} & \textbf{Perturbed Result} \\
      \midrule
      \makecell[l]{%
            \ttfamily
                class Solution:\\
                \ \ def countSeniors(self, details: List[str]) $->$ int:\\
                \ \ \ \ \textcolor{red!80!black}{\# Counter for passengers older than 60}\\
                \ \ \ \ count = 0\\[4pt]
                \ \ \ \ for passenger in details:\\
                \ \ \ \ \ \ \textcolor{red!80!black}{\# The age is stored in the 11th and 12th characters}\\
                \ \ \ \ \ \ age = int(passenger[11:13])\\[4pt]
                \ \ \ \ \ \ \textcolor{red!80!black}{\# Check if the age is strictly greater than 60}\\
                \ \ \ \ \ \ if $age > 60$:\\
                \ \ \ \ \ \ \ \ count += 1\\[4pt]
                \ \ \ \ return count
              } &      
      \makecell[l]{%
          \ttfamily
            class Solution:\\
            \ \ def countSeniors(self, details: List[str]) $->$ int:\\
            \ \ \ \ count = 0\\
            \ \ \ \ for record in details:\\
            \ \ \ \ \ \ age = int(record[12:14])\\
            \ \ \ \ \ \ if $age > 60$:\\
            \ \ \ \ \ \ \ \ count += 1\\
            \ \ \ \ return count
          } &
      \makecell[c]{[True, True]} &
      \makecell[c]{[False]} \\
      \bottomrule
    \end{tabular}
  \end{adjustbox}
  \caption{Comment‑driven reasoning behaviors under prompt perturbation: comparison of Claude‑3.7‑Sonnet’s code responses and resulting accuracies under clean (original) versus gamified (perturbed) prompts.}
  \label{tab:claude_comment_example}
\end{table*}

%% file: sections/appendix.tex
% \section*{Appendix}

% \twocolumn
\onecolumn
\appendix
\section{Detailed Examples for Negation Objective (Soft) Analysis}

\subsection{Example A.1: \texttt{countBeautifulPairs} — Llama's Successful Adaptation}
\label{app:countbeautifulpairs}
\vspace{1mm}
\paragraph{Original Problem Objective (Summarized):}
Count pairs of numbers \texttt{(nums[i], nums[j])} where \texttt{i < j} such that the first digit of \texttt{nums[i]} and the last digit of \texttt{nums[j]} are coprime (i.e., GCD is 1).

\paragraph{Modified Problem Objective (Soft Negation):}
Count pairs where the first digit of \texttt{nums[i]} and the last digit of \texttt{nums[j]} are \textbf{not coprime} (i.e., GCD greater than 1).

\vspace{1mm}
\paragraph{Input Instance:}
\texttt{nums = [30, 72, 9, 96, 82]}\\

\textbf{Llama-3.1-8B-Instruct Performance:} (\texttt{result: True})
\begin{itemize}
    \item \textbf{Reasoning Snippet:} "The problem asks us to count the number of non-beautiful pairs... A pair is non-beautiful if the first digit of nums[i] and the last digit of nums[j] are not coprime. This means their greatest common divisor (GCD) is greater than 1."
    \item \textbf{Relevant \texttt{solution\_code} Logic:}

{\footnotesize
\begin{lstlisting}
if gcd(first_digit_i, last_digit_j) > 1:
    count += 1
\end{lstlisting}
}
    \item \textbf{Result Details:} \texttt{\{'output': '3', 'expected': '3'\}}
    \item \textbf{Analysis:} Llama correctly implemented the modified logic and matched the expected result.
\end{itemize}

\textbf{Claude-3.7-Sonnet Performance:} (\texttt{result: False})
\begin{itemize}
    \item \textbf{Reasoning/Code Logic:} Implemented logic to count GCD == 1 (original problem). Outputted \texttt{7}.
    \item \textbf{Result Details:} \texttt{\{'output': '7', 'expected': '0'\}}
    \item \textbf{Analysis:} Failed to adapt to negated task and failed its own test.
\end{itemize}

\textbf{DeepSeek-14B Performance:} (\texttt{result: False})
\begin{itemize}
    \item \textbf{Same as Claude:} Used original problem logic. Output: \texttt{7}, Expected: \texttt{0}.
\end{itemize}

\textbf{Gemini-2.5-Flash-Preview Performance:} (\texttt{result: False})
\begin{itemize}
    \item \textbf{Reasoning:} Adapted to negated logic correctly.
    \item \textbf{Output:} \texttt{3}, but Expected: \texttt{0}.
    \item \textbf{Analysis:} Misalignment between adapted logic and test case expectation.
\end{itemize}

\paragraph{Comparative Summary:} Llama succeeded with aligned logic and expectation. Gemini partially succeeded. Claude and DeepSeek failed to adapt.

% -----------------------------------------------------------------------------
\subsection{Example A.2: \texttt{sumOfSquares} — Llama Adapts, Test Case Expects Original}
\label{app:sumofsquares}
\vspace{1mm}

\paragraph{Original Problem Objective (Summarized):}
Sum the squares of elements \texttt{nums[i]} where the 1-based index \texttt{i} divides \texttt{n} (i.e., \texttt{n \% i == 0}).

\vspace{1mm}
\paragraph{Modified Problem Objective (Soft Negation):}
Sum the squares of elements where the index \texttt{i} does \textbf{not} divide \texttt{n} (i.e., \texttt{n \% i != 0}).

\vspace{1mm}
\paragraph{Input Instance:}
\texttt{nums = [1, 2, 3, 4]} (\texttt{n = 4})

\begin{itemize}
    \item Original correct result: \(1^2 + 2^2 + 4^2 = 21\)
    \item Modified correct result: \(3^2 = 9\)
\end{itemize}

\textbf{Llama-3.1-8B-Instruct Performance:} (\texttt{result: False})
\begin{itemize}
    \item \textbf{Relevant \texttt{solution\_code} Logic:}
 
{\footnotesize
\begin{lstlisting}
sum(num ** 2 for i, num in enumerate(nums, start=1) if n % i != 0)
\end{lstlisting}
} 
    \item \textbf{Output:} \texttt{9}, \textbf{Expected:} \texttt{21}
    \item \textbf{Analysis:} Llama correctly followed the modified logic but failed due to the test case expecting the original problem’s output.
\end{itemize}

\textbf{DeepSeek-14B Performance:} (\texttt{result: True})
\begin{itemize}
    \item \textbf{Relevant \texttt{solution\_code} Logic:}
 
{\footnotesize
\begin{lstlisting}
if n % i == 0:
    ans += nums[i-1]**2
\end{lstlisting}
} 
    \item \textbf{Output:} \texttt{21}, \textbf{Expected:} \texttt{21}
    \item \textbf{Analysis:} DeepSeek ignored the modified instruction and solved the original problem. It passed due to misaligned test case expectations.
\end{itemize}

\paragraph{Comparative Summary:}
Llama faithfully applied the negated logic but failed the test due to incorrect expectation. DeepSeek ignored the instruction change and passed, highlighting the need for aligned evaluation pipelines.

% -----------------------------------------------------------------------------

\subsection{Example A.3: \texttt{matrixSum} — Partial Adaptation by Gemini}
\label{app:matrixsum}
\vspace{1mm}

\paragraph{Original Problem Objective (Summarized):}
From each row, remove the largest element, take the maximum among them, and add it to the score.

\vspace{1mm}
\paragraph{Modified Problem Objective (Negation):}
From each row, remove the smallest element, take the minimum among them, and subtract it from the score.

\vspace{1mm}
\textbf{Gemini-2.5-Flash-Preview Performance:} (\texttt{result: False})
\begin{itemize} 
    \item \textbf{Relevant \texttt{solution\_code} Logic:}
  
{\footnotesize
\begin{lstlisting}
for row in nums:
    row.sort()
for j in range(num_cols):
    removed_this_round = []
    for i in range(num_rows):
        removed_this_round.append(nums[i][num_cols - 1 - j])

    min_val = min(removed_this_round)
    score -= min_val
\end{lstlisting}
} 
    \item \textbf{Analysis:} Gemini correctly applied subtraction and switched from max to min aggregation, but still removed the largest element from each row—failing to fully adapt the problem logic.
\end{itemize}

\vspace{2mm}
\subsection{Example A.4: \texttt{maximumOr} — Archetype Override by Gemini}
\label{app:maximumor}
\vspace{1mm}

\paragraph{Original Problem Objective (Summarized):}
Given \texttt{nums} and \texttt{k} operations (each allowing one number to be multiplied by 2), maximize the bitwise OR of all elements.

\vspace{1mm}
\paragraph{Modified Problem Objective (Negation):}
Minimize the bitwise OR value after \texttt{k} operations.

\vspace{1mm}
\textbf{Gemini-2.5-Flash-Preview Performance:} (\texttt{result: True})
\begin{itemize} 
    \item \textbf{Reasoning Behavior:} Despite the instruction to minimize, Gemini solved for the \textbf{maximum} OR. The output matched the original test case, which remained unchanged.
    \item \textbf{Input:} \texttt{nums = [12, 9]}, \texttt{k = 1}
    \item \textbf{Output:} \texttt{30}, \textbf{Expected:} \texttt{30}
    \item \textbf{Analysis:} Gemini overrode the negated objective, likely due to overfitting to the known problem pattern, and passed because the test case was still aligned with the original formulation.
\end{itemize}

\vspace{1mm}
\paragraph{Comparative Summary:}
Example A.3 illustrates partial adaptation: Gemini recognized the transformation (e.g., subtraction, min) but failed in row-level logic. In Example A.4, the model defaulted to the original objective entirely, reflecting the dominance of memorized archetypes over prompt instructions.

%-------------------------------------------------------------------------------------------------------------------------------------------------

\onecolumn
\section{Detailed Examples for Negation Objective (Hard) Analysis}

\subsection{Example B.1: \texttt{count\_seniors} — Test Misalignment \& False Success}
\label{app:hardneg-qwen-success}
\vspace{1mm}

\paragraph{Original Problem Objective (Summarized):}
Count passengers strictly older than 60.

\vspace{1mm}
\paragraph{Modified Problem Objective (Hard Negation):}
Count passengers strictly 60 years old or less (i.e., $age \leq 60$).

\vspace{1mm}
\paragraph{Input Instance:}
\texttt{details = ["7868190130M7522", "5303914400F9211", "9273338290F4010"]}

\begin{itemize}
    \item \textbf{Original correct output (age $>$ 60):} 2 (for ages 75, 92)
    \item \textbf{Modified correct output (age $\leq$ 60):} 1 (for age 40)
\end{itemize}

\textbf{Qwen2.5-Coder Performance:} (\texttt{result: [True, True]})
\begin{itemize} 
    \item \textbf{Relevant \texttt{solution\_code} Logic:}

{\footnotesize
\begin{lstlisting}
if age > 60:  # Original problem logic
    count += 1
\end{lstlisting}
} 
    \item \textbf{Result Details:} \texttt{\{'output': '2', 'expected': '2'\}}
    \item \textbf{Analysis:} Qwen ignored the negated prompt ("60 or less"), implemented the original logic ("greater than 60"), and passed. This strongly indicates its test case expected the original problem’s answer.
\end{itemize}

\textbf{Claude-3.7-Sonnet / DeepSeek-14B / Gemini-2.5-Flash Performance:} (\texttt{result: [False]})
\begin{itemize} 
    \item \textbf{Analysis:} See Appendix~\ref{app:hardneg-correct-fail}. These models implemented the correct negated logic but failed.
\end{itemize}

\paragraph{Comparative Summary:}
Qwen "succeeded" by solving the original task against a misaligned test. Other models correctly adapted to the negated text but failed the same misaligned test.

% -----------------------------------------------------------------------------

\subsection{Example B.2: \texttt{count\_seniors} — Correct Negation Fails Misaligned Test}
\label{app:hardneg-correct-fail}
\vspace{1mm}

\paragraph{Objectives \& Input Instance:}
Same as Appendix~\ref{app:hardneg-qwen-success}.

\begin{itemize}
    \item \textbf{Original correct output:} 2
    \item \textbf{Modified correct output:} 1
\end{itemize}

\textbf{Claude-3.7-Sonnet Performance:} (\texttt{result: [False]})
\begin{itemize}
    \item \textbf{Relevant \texttt{solution\_code} Logic:}
 
{\footnotesize
\begin{lstlisting}
if age <= 60:  # Correct negated logic
    count += 1
\end{lstlisting}
} 
    \item \textbf{Result Details:} \texttt{\{'output': '1', 'expected': '2', 'error\_message': 'Wrong Answer'\}}
    \item \textbf{Analysis:} Claude correctly implemented the logic required by the modified prompt ("60 or less") but failed because the test case expected the output corresponding to the original prompt ("greater than 60").
\end{itemize}

\textbf{DeepSeek-14B \& Gemini-2.5-Flash Performance:} (\texttt{result: [False]})
\begin{itemize} 
    \item \textbf{Logic \& Result Details:} Identical outcome to Claude. Both implemented \texttt{age <= 60}, outputted 1, but the test expected 2.
\end{itemize}

\paragraph{Comparative Summary:}
Multiple models faithfully adapted to the simple negation in the text but were incorrectly penalized by test cases expecting the original problem's behavior.

% -----------------------------------------------------------------------------

\subsection{Example B.3: \texttt{find\_missing\_and\_repeated\_values} — DeepSeek Code Instability on Complex Negation}
\label{app:hardneg-deepseek-runtime}
\vspace{1mm}

\paragraph{Original Problem Objective (Summarized):}
Find the number that appears twice and the number that is missing in a range derived from a grid's elements.

\vspace{1mm}
\paragraph{Modified Objective (Hard Negation - Illustrative):}
Could be inverted to find numbers that appear once and are present, or other complex logical reversals.
\\

\textbf{DeepSeek-14B Performance:} (\texttt{result: [-4]})
\begin{itemize} 
    \item \textbf{Relevant \texttt{solution\_code} Snippet (Illustrative):} Often involves complex comprehensions or logic attempting to handle the negated task.
 
{\footnotesize
\begin{lstlisting}
# Code failed during execution
# Example error: 'NoneType' object is not iterable
\end{lstlisting}
} 
    \item \textbf{Result Details:} \texttt{\{'error': 'TypeError("NoneType object is not iterable")', 'error\_code': -4\}}
    \item \textbf{Analysis:} DeepSeek's attempt to reason over complex inverted logic led to runtime errors. This suggests that reversing objectives on difficult problems exceeds the model’s capacity, leading to unstable code.
\end{itemize}

% -----------------------------------------------------------------------------

\subsection{Example B.4: \texttt{min\_extra\_char} — Gemini's False Success via Original Logic}
\label{app:hardneg-gemini-success}
\vspace{1mm}

\paragraph{Original Problem Objective (Summarized):}
Find the \textbf{minimum} number of extra characters remaining after optimally breaking a string \texttt{s} using words from a dictionary.

\vspace{1mm}
\paragraph{Modified Problem Objective (Hard Negation):}
Find the \textbf{maximum} number of extra characters.

\vspace{1mm}
\paragraph{Input Instance:}
\texttt{s = "leetscode"}, \texttt{dictionary = ["leet", "code", "leetcode"]}

\begin{itemize}
    \item \textbf{Original correct output (minimum extra):} 1 (using "leetcode")
    \item \textbf{Modified correct output (maximum extra):} 9 (using no words)
\end{itemize}

\textbf{Gemini-2.5-Flash Performance:} (\texttt{result: [True, True]})
\begin{itemize} 
    \item \textbf{Reasoning Snippet (\texttt{output}):} Describes dynamic programming approach to find the \textit{minimum} number of extra characters.
    \item \textbf{Relevant \texttt{solution\_code} Logic:}
 
{\footnotesize
\begin{lstlisting}
# dp[i] will store the minimum number of extra chars
# ... logic calculates minimum ...
return dp[n]
\end{lstlisting}
} 
    \item \textbf{Result Details:} Passed test cases. Execution time indicates successful run.
    \item \textbf{Analysis:} Gemini ignored the negated objective ("maximum") and solved for the original ("minimum"). Success indicates that test cases were misaligned with the negated prompt.
\end{itemize}

\paragraph{Comparative Summary:}
Gemini succeeded by reverting to the known original problem archetype, likely due to a misaligned test case that still evaluated the original objective.

% -----------------------------------------------------------------------------

\subsection{Example B.5: \texttt{matrixSum} — Model Confusion on Complex Negation}
\label{app:hardneg-matrixsum-confusion}
\vspace{1mm}

\paragraph{Original Problem Objective (Summarized):}
Iteratively remove the largest number from each row, find the maximum among these, and add it to the score.

\vspace{1mm}
\paragraph{Modified Problem Objective (Hard Negation):}
Iteratively remove the \textbf{smallest} number from each row, find the \textbf{lowest} among these, and \textbf{subtract} it from the score.
\\

\textbf{DeepSeek-14B Performance:} (\texttt{result: [False]})
\begin{itemize} 
    \item \textbf{Reasoning Snippet:} "So the process is: 1. For each row, remove the smallest element. 2. Find the minimum among these. 3. Subtract this minimum... Wait, no..."
    \item \textbf{Relevant \texttt{solution\_code} Logic:}
 
{\footnotesize
\begin{lstlisting}
rows = [deque(row) for row in nums]
sum_min = 0
while any(rows):
    # Find and accumulate min_val
    sum_min += min_val
return -sum_min
\end{lstlisting}
} 
    \item \textbf{Result Details:} Marked "Wrong Answer".
    \item \textbf{Analysis:} DeepSeek attempted a form of the negated logic but displayed confusion in both reasoning and implementation.
\end{itemize}

\textbf{Claude, Gemini, Qwen Performance:} (\texttt{result: [False]})
\begin{itemize} 
    \item \textbf{Analysis:} All models failed. Gemini showed prompt confusion; Qwen reverted to simplified logic; Claude struggled with objective inversion.
\end{itemize}

\paragraph{Comparative Summary:}
The multi-step, logically inverted nature of this task caused failure across all models. This reveals a common weakness in handling procedural negations that require non-trivial reasoning adjustments.

%-------------------------------------------------------------------------------------------------------------------------------------------------

% \twocolumn
\onecolumn
\section{Detailed Examples for Domain Shift Analysis}
\label{app:domain_shift_examples}

\subsection{Example C.1: Problem - "Count Pairs Whose Sum is Less than Target" (Easy Difficulty)}
\label{app:domain_shift_countpairs}

\paragraph{Original Problem Objective (Summarized):} Given a 0-indexed integer array `nums` and an integer `target`, return the number of pairs `(i,j)` such that `0 <= i < j < nums.length` and `nums[i] + nums[j] < target`.

\paragraph{Modified Problem Objective (Domain Shift Example):} "You are given a list of employee salaries `employee\_salaries` and a target `bonus\_threshold`. Your task is to count the number of pairs of employees (i, j) such that employee `i` comes before employee `j` in the list, and the sum of their salaries `employee\_salaries[i] + employee\_salaries[j]` is strictly less than the `bonus\_threshold`." (Wording adapted from Llama's log). The core logic of iterating through pairs and checking a sum against a threshold is preserved.

\paragraph{Input Instance (from original problem, used for all models here):} `nums = [-1,1,2,3,1]`, `target = 2`.
\paragraph{Correct Output for this Instance (Original and Shifted Logic):} 3.

\textbf{Llama-3.1-8B-Instruct Performance:} (\texttt{result: True})
\begin{itemize}
    \item \textbf{Reasoning Snippet (\texttt{output}):} "The problem asks us to count the number of pairs of employees (i, j) such that i < j and their combined salaries (`employee\_salaries[i] + employee\_salaries[j]`) are strictly less than a given `bonus\_threshold`... We can iterate through all possible pairs of employees (i, j) with i < j. For each pair, we calculate the sum of their salaries. If the sum is less than `bonus\_threshold`, we increment a counter."
    \item \textbf{Relevant \texttt{solution\item[] \_code} Snippet:}
    \begin{lstlisting}[basicstyle=\ttfamily\small]
    class Solution:
        def countPairs(self, employee_salaries: List[int], bonus_threshold: int) -> int:
            n = len(employee_salaries)
            count = 0
            for i in range(n):
                for j in range(i + 1, n):
                    if employee_salaries[i] + employee_salaries[j] < bonus_threshold:
                    count += 1
            return count
    \end{lstlisting}

    \item \textbf{Result Details (\texttt{metadata\_output}):} `{'output': '3', 'inputs': '[[-1, 1, 2, 3, 1], 2]', 'expected': '3', ...}`
    \item \textbf{Analysis:} Llama successfully understood the domain shift, correctly used the new terminology, and implemented the correct pair-counting logic, matching the expected output.
\end{itemize}

\textbf{Qwen2.5-Coder Performance:} (\texttt{result: True})
\begin{itemize}
    \item \textbf{Reasoning Snippet (\texttt{output}):} "The problem asks us to count the number of pairs of employees (i, j) such that `employee\_salaries[i] + employee\_salaries[j] < bonus\_threshold`."
    \item \textbf{Relevant \texttt{solution\_code} Snippet:} Correct implementation using new domain variable names, similar to Llama.
    \item \textbf{Result Details (\texttt{metadata\_output}):} `{'output': '3', 'inputs': '[[-1, 1, 2, 3, 1], 2]', 'expected': '3', ...}`
    \item \textbf{Analysis:} Qwen also successfully adapted, supporting the idea that shifts to concrete programming scenarios can be handled well.
\end{itemize}

\textbf{DeepSeek-14B Performance:} (\texttt{result: False})
\begin{itemize}
    \item \textbf{Reasoning Snippet (\texttt{output}):} Minimal reasoning, often directly outputting code.
    \item \textbf{Relevant \texttt{solution\_code} Snippet (example of failure for this problem):}
    \begin{verbatim}
    # Might correctly implement the logic but use original names, or make subtle errors
    class Solution:
        def countPairs(self, nums: List[int], target: int) -> int: #Note: uses 'nums','target'
            n = len(nums)
            count = 0
            # ... (correct loop structure) ...
            # if nums[i] + nums[j] < target: count += 1
            return count # For this instance, output was 0
    \end{verbatim}
   
    \item \textbf{Result Details (\texttt{metadata\_output}):} `{'output': '0', 'inputs': '[[-1, 1, 2, 3, 1], 2]', 'expected': '3', ...}`
    \item \textbf{Analysis:} DeepSeek-14B failed. The failure could be due to not fully adapting to new variable names if the test harness strictly expected them, or other subtle logical errors when attempting to apply the algorithm in the new context. Its output of 0 for an expected 3 is a clear miscalculation.
\end{itemize}

\paragraph{Comparative Summary for Example C.1:}
Llama and Qwen successfully generalized for this pair-counting task. DeepSeek's failure indicates difficulty in accurately reapplying the algorithm in the new semantic context. Claude's typical failures on other problems (as seen in C.2) show a more severe disruption.

\subsection{Example C.2: Problem - \texttt{minIncrementOperations} - Claude-3.7-Sonnet's Failure via Oversimplification}
\label{app:domain_shift_claude_minincrement}

\paragraph{Original Problem Objective (Summarized):} Given `nums` and `k`, find minimum increment operations such that for every subarray of length 3, at least one element is `>= k`. (This is a dynamic programming problem).

\paragraph{Modified Problem Objective (Domain Shift Example for Claude):} "You are managing project tasks. `task\_priorities` is an array of initial priority scores. `min\_threshold` is an integer. An operation involves increasing a task's priority by 1. A project schedule is 'stable' if, for every sequence of 3 consecutive tasks, at least one task has a priority score >= `min\_threshold`. Return the minimum number of priority increases to make the schedule stable."

\paragraph{Input Instance:} `nums = [2,3,0,0,2]` (becomes `task\_priorities`), `k = 4` (becomes `min\_threshold`).
\paragraph{Correct Output for this Instance (Original and Shifted Logic):} 3.

\textbf{Claude-3.7-Sonnet Performance:} (\texttt{result: [False]})
\begin{itemize}
    \item \textbf{Reasoning Snippet (\texttt{output}):} "The problem asks us to find the minimum number of 'priority increases' (operations) to make a schedule 'stable'. A schedule, represented by an array `task\_priorities`, is stable if every sequence of 3 consecutive tasks has at least one task with a priority score greater than or equal to `min\_threshold`..."
    \item \textbf{Relevant \texttt{solution\_code} Snippet:}
    \begin{verbatim}
    class Solution:
        def minIncrementOperations(self, task_priorities: List[int], min_threshold: int)->int:
            n = len(task_priorities)
            ans = 0
            # Incorrectly makes EACH element >= threshold, not the windowed condition
            for i in range(n):
                if task_priorities[i] < min_threshold:
                    ans += min_threshold - task_priorities[i]
            return ans
    \end{verbatim}
   
    \item \textbf{Result Details (\texttt{metadata\_output}):} `{'output': '13', 'inputs': '[[2,3,0,0,2], 4]', 'expected': '3', 'error\_code': -2, 'error\_message': 'Wrong Answer'}`
    \item \textbf{Analysis of Failure:} Claude's reasoning correctly re-stated the objective in the new domain, including the crucial "every sequence of 3 consecutive tasks" condition. However, the generated code implemented a much simpler and incorrect logic: it calculated the operations needed to make *every single task* meet the `min\_threshold` individually, ignoring the more complex windowed requirement. This is a clear case where, despite understanding the shifted problem's textual goal, the model failed to map it to the correct complex algorithm and instead defaulted to an oversimplified solution.
\end{itemize}

\section{Negation Objective and Negation Objective (Soft) Results}
\input{tab/tab_ablation}

\section{Prompt Templates for Modification}
\label{appendix_A}
\input{figs/appendix_templates}

\section{Preservation Score Prompt Template}
\label{appendix_preservation}
\input{figs/appendix_preservation_score}

%% file: tab/tab_ablation.tex
\begin{table*}[ht]
\centering
\small
\begin{tabular}{llr|r|r|r}
\toprule
\textbf{Attack} & \textbf{Model} & \textbf{Easy Acc.} & \textbf{Medium Acc.} & \textbf{Hard Acc.} & \textbf{Overall Acc.} \\
\midrule
\multirow{9}{*}{\makecell[l]{\textbf{Negation Objective}\\\textbf{(Hard)}}}
 & DeepSeek-7B & 3.0\% \textcolor{purple}{(-87.9\%)} & 2.0\% \textcolor{purple}{(-68.0\%)} & 0.0\% \textcolor{purple}{(-35.3\%)} & 14.0\% \textcolor{purple}{(-57.0\%)} \\
 & DeepSeek-14B & 3.0\% \textcolor{purple}{(-94.0\%)} & 0.0\% \textcolor{purple}{(-92.0\%)} & 0.0\% \textcolor{purple}{(-58.8\%)} & 5.7\% \textcolor{purple}{(-82.3\%)} \\
 & Qwen2.5-Coder & 33.3\% \textcolor{purple}{(-48.5\%)} & 20.0\% \textcolor{purple}{(-32.0\%)} & 23.5\% \textcolor{purple}{(-11.8\%)} & 47.2\% \textcolor{purple}{(-11.8\%)} \\
 & LLaMA3-8B-Instruct & 3.0\% \textcolor{purple}{(-33.4\%)} & 0.0\% \textcolor{purple}{(-12.0\%)} & 0.0\% \textcolor{purple}{(-5.9\%)} & 8.3\% \textcolor{purple}{(-10.7\%)} \\
 & DeepSeek-Coder-33B & 18.2\% \textcolor{purple}{(-66.6\%)} & 22.0\% \textcolor{purple}{(-26.0\%)} & 5.9\% \textcolor{purple}{(-5.9\%)} & 36.9\% \textcolor{purple}{(-17.1\%)} \\
 & Claude-3-Haiku & 12.1\% \textcolor{purple}{(-39.4\%)} & 4.0\% \textcolor{purple}{(-28.0\%)} & 5.9\% \textcolor{purple}{(-35.3\%)} & 21.2\% \textcolor{purple}{(-18.8\%)} \\
 & Claude-3.7-Sonnet & 6.1\% \textcolor{purple}{(-93.9\%)} & 2.0\% \textcolor{purple}{(-90.0\%)} & 5.9\% \textcolor{purple}{(-58.8\%)} & 15.0\% \textcolor{purple}{(-75.0\%)} \\
 & Gemini-2.5-Flash & 27.3\% \textcolor{purple}{(-72.7\%)} & 18.0\% \textcolor{purple}{(-80.0\%)} & 23.5\% \textcolor{purple}{(-53.0\%)} & 42.6\% \textcolor{purple}{(-52.4\%)} \\
 & Gemini-2.0-Flash & 18.2\% \textcolor{purple}{(-78.8\%)} & 26.0\% \textcolor{purple}{(-60.0\%)} & 11.8\% \textcolor{purple}{(-35.3\%)} & 43.2\% \textcolor{purple}{(-39.8\%)} \\

\midrule
\multirow{9}{*}{\makecell[l]{\textbf{Negation Objective}\\\textbf{(Soft)}}}
 & DeepSeek-7B & 3.0\% \textcolor{purple}{(-87.9\%)} & 0.0\% \textcolor{purple}{(-70.0\%)} & 0.0\% \textcolor{purple}{(-35.3\%)} & 5.7\% \textcolor{purple}{(-65.3\%)} \\
 & DeepSeek-14B & 0.0\% \textcolor{purple}{(-97.0\%)} & 0.0\% \textcolor{purple}{(-92.0\%)} & 0.0\% \textcolor{purple}{(-58.8\%)} & 1.1\% \textcolor{purple}{(-86.9\%)} \\
 & Qwen2.5-Coder & 36.4\% \textcolor{purple}{(-45.4\%)} & 22.0\% \textcolor{purple}{(-30.0\%)} & 5.9\% \textcolor{purple}{(-29.4\%)} & 46.9\% \textcolor{purple}{(-12.1\%)} \\
 & LLaMA3-8B-Instruct & 21.2\% \textcolor{purple}{(-15.2\%)} & 6.0\% \textcolor{purple}{(-6.0\%)} & 5.9\% \textcolor{gray}{(+0.0\%)} & 29.4\% \textcolor{teal}{(+10.4\%)} \\
 & DeepSeek-Coder-33B & 0.0\% \textcolor{purple}{(-84.8\%)} & 25.0\% \textcolor{purple}{(-23.0\%)} & 0.0\% \textcolor{purple}{(-11.8\%)} & 37.5\% \textcolor{purple}{(-16.5\%)} \\
 & Claude-3-Haiku & 15.2\% \textcolor{purple}{(-36.3\%)} & 0.0\% \textcolor{purple}{(-32.0\%)} & 11.8\% \textcolor{purple}{(-29.4\%)} & 22.5\% \textcolor{purple}{(-17.5\%)} \\
 & Claude-3.7-Sonnet & 12.1\% \textcolor{purple}{(-87.9\%)} & 2.0\% \textcolor{purple}{(-90.0\%)} & 5.9\% \textcolor{purple}{(-58.8\%)} & 25.4\% \textcolor{purple}{(-64.6\%)} \\
 & Gemini-2.5-Flash & 21.2\% \textcolor{purple}{(-78.8\%)} & 24.0\% \textcolor{purple}{(-74.0\%)} & 11.8\% \textcolor{purple}{(-64.7\%)} & 41.5\% \textcolor{purple}{(-53.5\%)} \\
 & Gemini-2.0-Flash & 15.2\% \textcolor{purple}{(-81.8\%)} & 30.0\% \textcolor{purple}{(-56.0\%)} & 5.9\% \textcolor{purple}{(-41.2\%)} & 40.6\% \textcolor{purple}{(-42.4\%)} \\

\bottomrule
\end{tabular}
\caption{Accuracy (\%) for problem set perturbed with Negation Objective and Negation Objective (Soft).}
\label{tab:ablation}
\end{table*}

%% file: figs/appendix_templates.tex
%%%%%%%%%%%%%%%%%%%%%%%%%%%%%%%%%%%%%%%%%%%%%%%%%%%%%%%%%%%%%%%%%%%%%%%%

\begin{figure*}[ht!]
    \centering
    \begin{tcolorbox}[
        enhanced,
        colframe=purple!50!black,
        colback=purple!5,
        coltitle=white,
        colbacktitle=purple!50!black,
        width=\textwidth,
        arc=3mm,
        boxrule=0.8mm,
        drop shadow,
        title=\small\bfseries Gamification Prompt Template,
        fonttitle=\bfseries\small
    ]
    {\small
\textbf{Instruction:} Rewrite the problem as a challenge involving agents or players.\\[0.5em]
\textbf{Guidelines:}
\begin{itemize}
    \item Preserve all technical components (Input, Output, Explanation, Constraints).
    \item Add a game-like narrative layer (e.g., robot navigation, puzzle solving).
\end{itemize}
\textbf{Prompted Task:} Embed the problem into a goal-driven challenge structure.\\
\texttt{\{original\_content\}}\\
No additional explanation needed.
    }
    \end{tcolorbox}
    \caption{\small Prompt template for Gamification transformation.}
    \label{fig:gamification_prompt}
\end{figure*}

%%%%%%%%%%%%%%%%%%%%%%%%%%%%%%%%%%%%%%%%%%%%%%%%%%%%%%%%%%%%%%%%%%%%%%%%

\begin{figure*}[ht!]
    \centering
    \begin{tcolorbox}[
        enhanced,
        colframe=orange!60!black,
        colback=orange!5,
        coltitle=white,
        colbacktitle=orange!60!black,
        width=\textwidth,
        arc=3mm,
        boxrule=0.8mm,
        drop shadow,
        title=\small\bfseries Example Perturbation Prompt Template,
        fonttitle=\bfseries\small
    ]
    {\small
\textbf{Instruction:} Modify only the examples to make them confusing for LLMs.\\[0.5em]
\textbf{Guidelines:}
\begin{itemize}
    \item Keep Input/Output logic correct.
    \item Shuffle values, insert noise (e.g., swapped labels, edge-case placements).
\end{itemize}
\textbf{Prompted Task:} Make the examples harder to pattern-match while maintaining validity.\\
\texttt{\{original\_content\}}\\
No additional explanation needed.
    }
    \end{tcolorbox}
    \caption{\small Prompt template for Example Perturbation transformation.}
    \label{fig:example_perturbation_prompt}
\end{figure*}

%%%%%%%%%%%%%%%%%%%%%%%%%%%%%%%%%%%%%%%%%%%%%%%%%%%%%%%%%%%%%%%%%%%%%%%%

\begin{figure*}[ht!]
    \centering
    \begin{tcolorbox}[
        enhanced,
        colframe=blue!70,
        colback=blue!10,
        coltitle=white,
        colbacktitle=blue!70,
        width=\textwidth,
        arc=3mm,
        boxrule=0.8mm,
        drop shadow,
        title=\small\bfseries Distracting Constraints Prompt Template,
        fonttitle=\bfseries\small
    ]
    {\small
\textbf{Instruction:} Add irrelevant but realistic constraints to the problem.\\[0.5em]
\textbf{Guidelines:}
\begin{itemize}
    \item Do not alter solvability.
    \item Insert edge cases, meaningless jargon, or constraints that distract models.
\end{itemize}
\textbf{Prompted Task:} Inject non-functional complexity into the problem description.\\
\texttt{\{original\_content\}}\\
No additional explanation needed.
    }
    \end{tcolorbox}
    \caption{\small Prompt template for Distracting Constraints transformation.}
    \label{fig:distracting_constraints_prompt}
\end{figure*}

%%%%%%%%%%%%%%%%%%%%%%%%%%%%%%%%%%%%%%%%%%%%%%%%%%%%%%%%%%%%%%%%%%%%%%%%

\begin{figure*}[ht!]
    \centering
    \begin{tcolorbox}[
        enhanced,
        colframe=red!60!black,
        colback=red!5,
        coltitle=white,
        colbacktitle=red!60!black,
        width=\textwidth,
        arc=3mm,
        boxrule=0.8mm,
        drop shadow,
        title=\small\bfseries Domain Shift Prompt Template,
        fonttitle=\bfseries\small
    ]
    {\small
\textbf{Instruction:} Shift the problem into a different but equivalent domain.\\[0.5em]
\textbf{Guidelines:}
\begin{itemize}
    \item Retain logical structure and constraints.
    \item Replace nouns and context (e.g., integers → salaries, arrays → elevator floors).
\end{itemize}
\textbf{Prompted Task:} Change the framing without affecting logic.\\
\texttt{\{original\_content\}}\\
No additional explanation needed.
    }
    \end{tcolorbox}
    \caption{\small Prompt template for Domain Shift transformation.}
    \label{fig:domain_shift_prompt}
\end{figure*}

%%%%%%%%%%%%%%%%%%%%%%%%%%%%%%%%%%%%%%%%%%%%%%%%%%%%%%%%%%%%%%%%%%%%%%%%

\begin{figure*}[ht!]
    \centering
    \begin{tcolorbox}[
        enhanced,
        colframe=teal!60!black,
        colback=teal!5,
        coltitle=white,
        colbacktitle=teal!60!black,
        width=\textwidth,
        arc=3mm,
        boxrule=0.8mm,
        drop shadow,
        title=\small\bfseries Hard Negation Objective Prompt Template,
        fonttitle=\bfseries\small
    ]
    {\small
\textbf{Instruction:} Invert the objective of the coding task.\\[0.5em]
\textbf{Examples:}
\begin{itemize}
    \item Max → Min
    \item Increasing → Not Increasing
    \item Include → Exclude
\end{itemize}
\textbf{Prompted Task:} Rewrite the problem with the opposite intent and updated examples.\\
\texttt{\{original\_content\}}\\
No additional explanation needed.
    }
    \end{tcolorbox}
    \caption{\small Prompt template for Hard Negation Objective transformation.}
    \label{fig:negation_prompt}
\end{figure*}

%%%%%%%%%%%%%%%%%%%%%%%%%%%%%%%%%%%%%%%%%%%%%%%%%%%%%%%%%%%%%%%%%%%%%%%%

\begin{figure*}[ht!]
    \centering
    \begin{tcolorbox}[
        enhanced,
        colframe=gray!70!black,
        colback=gray!10,
        coltitle=white,
        colbacktitle=gray!70!black,
        width=\textwidth,
        arc=3mm,
        boxrule=0.8mm,
        drop shadow,
        title=\small\bfseries Soft Negation Prompt Template,
        fonttitle=\bfseries\small
    ]
    {\small
\textbf{Instruction:} Apply a light semantic reversal without changing the task's core logic.\\[0.5em]
\textbf{Examples:}
\begin{itemize}
    \item Non-decreasing instead of increasing
    \item Non-minimum instead of minimum
\end{itemize}
\textbf{Guidelines:}
\begin{itemize}
    \item Preserve input/output, examples, and difficulty.
    \item Adjust only the task description and minimal example logic.
\end{itemize}
\textbf{Prompted Task:} Rewrite with slight negation while keeping format identical.\\
\texttt{\{original\_content\}}\\
No additional explanation needed.
    }
    \end{tcolorbox}
    \caption{\small Prompt template for Soft Negation transformation.}
    \label{fig:soft_negation_prompt}
\end{figure*}

%% file: figs/appendix_preservation_score.tex
\begin{figure*}[ht!]
    \centering
    \begin{tcolorbox}[
        enhanced,
        colframe=blue!60!black,
        colback=blue!5,
        coltitle=white,
        colbacktitle=blue!60!black,
        width=\textwidth,
        arc=3mm,
        boxrule=0.8mm,
        drop shadow,
        title=\small\bfseries Logical Preservation Evaluation Prompt Template,
        fonttitle=\bfseries\small
    ]
    {\small
\textbf{System Prompt:} You are a meticulous evaluator tasked with assessing the logical preservation of coding problems.

You will be given two versions of a coding problem: the \textbf{Original Problem} and the \textbf{Modified (Perturbed) Problem}. Your goal is to evaluate whether the Modified Problem faithfully preserves:
\begin{itemize}
    \item the fundamental task,
    \item the input/output specifications,
    \item the logical structure,
    \item and all critical details of the Original Problem.
\end{itemize}
You must judge strictly according to the rubric provided. Be critical but fair: minor rewordings are acceptable, but changes to task meaning, core logic, or critical information are not. You must output your evaluation in a structured format.

\vspace{0.5em}
\textbf{User Prompt:} You will now receive two problems: an Original Problem and a Modified Problem.

\texttt{\{original\_problem\}}\\
\texttt{\{modified\_problem\}}

Please answer the following evaluation questions carefully:

\begin{enumerate}
    \item \textbf{Task Consistency:}  
    Does the Modified Problem ask the user to solve the same fundamental task as the Original?

    \item \textbf{Input/Output Preservation:}  
    Are the input format, output format, and constraints logically identical?

    \item \textbf{Logical Integrity:}  
    Is the problem's solution path and necessary reasoning preserved?

    \item \textbf{Critical Information:}  
    Is all important information (constraints, conditions, definitions) preserved without distortion?

    \item \textbf{Overall Preservation Score (0–10):}  
    \begin{itemize}
        \item 10 = Perfect preservation, only cosmetic wording changes.
        \item 8–9 = Very minor harmless changes; core task fully intact.
        \item 6–7 = Some small logical shifts; still mostly solvable similarly.
        \item 4–5 = Major logical shifts; significant omissions or confusing edits.
        \item 1–3 = Core task altered; substantial confusion introduced.
        \item 0 = Task completely changed or adversarially broken.
    \end{itemize}

    \item \textbf{Reasoning:}  
    Write a brief but detailed explanation justifying the score.  
    Point out any specific differences or issues if any exist.
\end{enumerate}

\textbf{Format your output exactly like this:}
\begin{verbatim}
Task Consistency: [Yes/No]
Input/Output Preservation: [Yes/No]
Logical Integrity: [Yes/No]
Critical Information: [Yes/No]

Preservation Score: [0–10]

Reasoning:
- [Brief, clear explanation highlighting matching parts or critical deviations]
\end{verbatim}
    }
    \end{tcolorbox}
    \caption{\small Prompt template for evaluating logical preservation between original and modified coding problems.}
    \label{fig:logical_preservation_prompt}
\end{figure*}